\documentclass[pmlr]{jmlr}

\makeatletter
\let\JMLR@endabstract\endabstract
\makeatother

\RequirePackage{graphicx}

\usepackage{arabtex}
\AtBeginDocument{\setcode{utf8}}
\newcommand{\arb}[1]{\begingroup\setcode{utf8}\RL{#1}\endgroup}

\makeatletter
\AtBeginDocument{\let\endabstract\JMLR@endabstract}
\makeatother

\usepackage{booktabs}
\usepackage{longtable}
\usepackage{multirow}    
\usepackage{xcolor}
\usepackage{array}      
 \usepackage{float}
\usepackage{makecell}
\usepackage{siunitx}
\usepackage[T1]{fontenc}
\usepackage{caption}

\usepackage{etoolbox}
\makeatletter
\patchcmd{\section}{\@afterindenttrue}{\@afterindentfalse}{}{}
\patchcmd{\subsection}{\@afterindenttrue}{\@afterindentfalse}{}{}
\makeatother

\theorembodyfont{\upshape}
\theoremheaderfont{\scshape}
\theorempostheader{:}
\theoremsep{\newline}

\jmlrvolume{298} 
\jmlrpages{1-37}
\jmlryear{2025}
\jmlrworkshop{Machine Learning for Healthcare}

\title[MedArabiQ: Benchmarking LLMs on Arabic Medical Tasks]{MedArabiQ: Benchmarking Large Language Models on Arabic Medical Tasks}

\newcommand{\authdag}{\textsuperscript{\textdagger}}

\author{%
\Name{Mouath Abu Daoud}\authdag\  \Email{mma9138@nyu.edu}\\
\Name{Chaimae Abouzahir}\authdag\  \Email{ca2627@nyu.edu}\\
\Name{Leen Kharouf} \Email{lk2713@nyu.edu}\\
\Name{Walid Al-Eisawi} \Email{wa2076@nyu.edu}\\
\Name{Nizar Habash} \Email{nizar.habash@nyu.edu}\\
\Name{Farah E. Shamout} \Email{farah.shamout@nyu.edu}\\
\addr {New York University Abu Dhabi, United Arab Emirates}\\
\addr {\authdag\ Equal contributions}
}

\begin{document}

\maketitle

\begin{abstract}
Large Language Models (LLMs) have demonstrated significant promise for various applications in healthcare. However, their effectiveness in the Arabic medical domain remains unexplored due to the lack of high-quality domain-specific datasets and benchmarks. This study introduces MedArabiQ, a new benchmark dataset consisting of seven Arabic medical tasks, covering multiple specialties and including multiple-choice questions, fill-in-the-blank questions, and patient-doctor questions and answers. We first constructed the dataset using past medical exams as well as publicly available datasets. We conducted an extensive evaluation with eight state-of-the-art open-access and proprietary high-resource LLMs, including GPT-4, Deepseek v3, and Gemini 1.5. Our findings highlight the need for the creation of new high-quality benchmarks that span different languages to ensure fair deployment and scalability of LLMs in healthcare. By establishing this benchmark and releasing the dataset, we provide a foundation for future research aimed at evaluating and enhancing the multilingual capabilities of LLMs for the equitable use of generative AI in healthcare.
\end{abstract}

\paragraph*{Data Availability}
In this article, we present a new benchmark dataset, MedArabiQ, designed to evaluate the performance of LLMs on Arabic medical tasks. We release our data for reproducibility to enable a fair evaluation of language models in the future: \url{https://github.com/nyuad-cai/MedArabiQ}

\section{Introduction}
\label{sec: introduction}
The recent advent of Large Language Models (LLMs) has revolutionized Natural Language Processing (NLP) by demonstrating exceptional performance in various tasks, ranging from language translation to creative writing \citep{mdpi2023}. Although LLMs were initially designed for general language understanding, they have since been evaluated for many domain-specific applications, such as education, programming, art, and medicine. They have also been adapted for domain-specific tasks using different fine-tuning strategies and specialized datasets \citep{arxiv2023}. 

The use of LLMs in healthcare has sparked enthusiasm due to the potential to improve diagnostic processes, clinical decision-making, and overall patient care \citep{mdpi2023,sciencedirect2024}. One particular application of interest is medical education, in which LLMs can generate concise summaries of information and support interactive learning experiences \citep{oup2023}. To this end, several benchmarks have been proposed to assess the capabilities of LLMs in medical knowledge and reasoning. Despite these advances, challenges remain, including ethical concerns, risks of generating biased or harmful content, and variability in performance in different languages and cultural contexts \citep{wiley2024, mdpi2023}.

Existing benchmarks, such as GLUE and MedQA, cater primarily to English, leaving a significant gap in the evaluation of LLMs for Arabic healthcare applications \citep{nature2024}. This is due to multiple factors, including the limited availability of high-quality Arabic datasets for clinical applications, coupled with the unique linguistic complexity of Arabic, especially when considering the many different Arabic dialect regions (e.g., Gulf, Maghreb, Egypt, Levant, among others) in addition to Modern Standard Arabic (MSA) \citep{acl2018}. Additionally, although various multilingual models include Arabic in their training data, their performance often falls short in clinical contexts due to insufficient domain-specific resources and a lack of appropriate benchmarks \citep{mdpi2023, arxiv2024}.  Addressing these gaps is critical to unlocking the full potential of LLMs for Arabic-speaking patients and providers, ensuring equitable access to AI-driven advancements in healthcare.

To address these gaps, there is a growing need for frameworks that assess  LLM performance in clinical tasks specific to Arabic-speaking populations. By developing benchmarks that reflect real-world clinical interactions, we can ensure more reliable and culturally appropriate LLM deployment in multilingual healthcare systems.  
In this study, we make several key contributions to address these challenges by introducing MedArabiQ (see Figure~\ref{fig:overview}). \color{black} First, we developed seven benchmark datasets designed to evaluate LLMs in Arabic healthcare applications while addressing Arabic linguistic complexity along with domain-specific challenges that hinder existing models. We focus on critical healthcare medical tasks, including medical question-answering, clinical dialogue, and ethical decision making. Secondly, we analyze the performance of multilingual and Arabic LLMs, including high-resource proprietary and open-access LLMs, highlighting the impact of linguistic coverage and transparency in training data on healthcare applications. We conduct a comprehensive evaluation to assess model performance, in order to provide a robust foundation for advancing AI-driven solutions for Arabic healthcare tasks.

\subsection*{Generalizable Insights about Machine Learning in the Context of Healthcare}
Our work provides a strong foundation for advancing LLM evaluation for Arabic medical tasks. While our work focuses on MSA, future efforts could further develop our preprocessing pipeline to explore the use of regional dialects, which are commonly used in clinical communication. MedArabiQ also eliminates the entry barrier for researchers in underrepresented regions by providing annotated data and an evaluation framework. By promoting open data and reproducibility, we believe that this is an important step towards building LLM agents that are more inclusive and globally relevant. Overall, MedArabiQ supports further research in areas like clinical decision support and telehealth, ultimately contributing to the development of AI models that are better suited for broader adoption in global real-world clinical workflows.




\section{Related Work}
Several benchmarks have been proposed for the evaluation of LLMs for various medical tasks, predominantly in the English language. For example, \cite{GAO2023104286} introduce Dr. Bench, a diagnostic reasoning benchmark for clinical NLP emphasizing clinical text understanding, medical knowledge reasoning, and diagnosis generation. The benchmark combines English-language data from various sources but primarily consists of in-hospital clinical notes. To assess LLM performance in medical question-answering tasks, a number of benchmarks incorporated samples from medical board exams, such as MedQA. MedQA advances multilingual benchmarking by including questions in traditional and simplified Chinese and English \citep{jin2020diseasedoespatienthave}. The Massive Multitask Language Understanding (MMLU) benchmark uses the US Medical Licensing Exam (USMLE) \citep{hendrycks2021measuringmassivemultitasklanguage} for a subset of tasks. MedMCQA extends these benchmarks to a multilingual evaluation framework \citep{pal2022medmcqa}.

Despite the growing interest in Arabic NLP, Arabic medical benchmarks remain limited. \cite{achiam2023gpt} translated MMLU into 14 languages, including Arabic, with the help of professional human translators. AraSTEM focuses on the task of question-answering and includes a medical subset \citep{mustapha2024arastemnativearabicmultiple}. AraMed similarly presents an Arabic medical corpus and an annotated Arabic question-answering dataset \citep{alasmari2024}. Other datasets are also largely skewed towards the task of Arabic medical question-answering and have other limitations, summarized in Table \ref{table:dataset_comparison}. While these resources are helpful, they do not comprehensively cover the spectrum of Arabic medical tasks, highlighting the need for dedicated benchmarking efforts. 

Clinically, a number of frameworks have been proposed for evaluating clinical AI models. The ‘Governance Model for AI in Healthcare’ consists of four main pillars: fairness, transparency, trustworthiness, and accountability \citep{Reddy2020491}. Similarly, \cite{dada2024doesbiomedicaltraininglead} propose the Clinical Language Understanding Evaluation framework, which was designed to assess LLMs with real patient data for various modalities, including patient-specific question-answering, hypothesis deduction, and problem and inquiry summarization. \cite{kanithi2024} introduce ‘MEDIC’, a framework for evaluating LLMs across five clinically relevant categories: medical reasoning, ethical and bias concerns, data and language understanding, in-context learning, and clinical safety and risk assessment. Although there has been significant progress in the field of benchmarking LLMs in medical tasks, most evaluations have been performed on data in English, and there is no single benchmark to assess LLMs in Arabic on more than one medical task and performance metric. There are over 380 million native Arabic speakers \citep{Eberhard2024}, a number of which are monolingual. With the vast potential of LLMs in healthcare, it is crucial to accommodate Arabic-speaking patients to ensure fair deployment. Further details on related works are present in Appendix \ref{appendix-a}.

\section{Methods}
\begin{figure}[t!]
    \centering
    \includegraphics[width=1.0\textwidth]{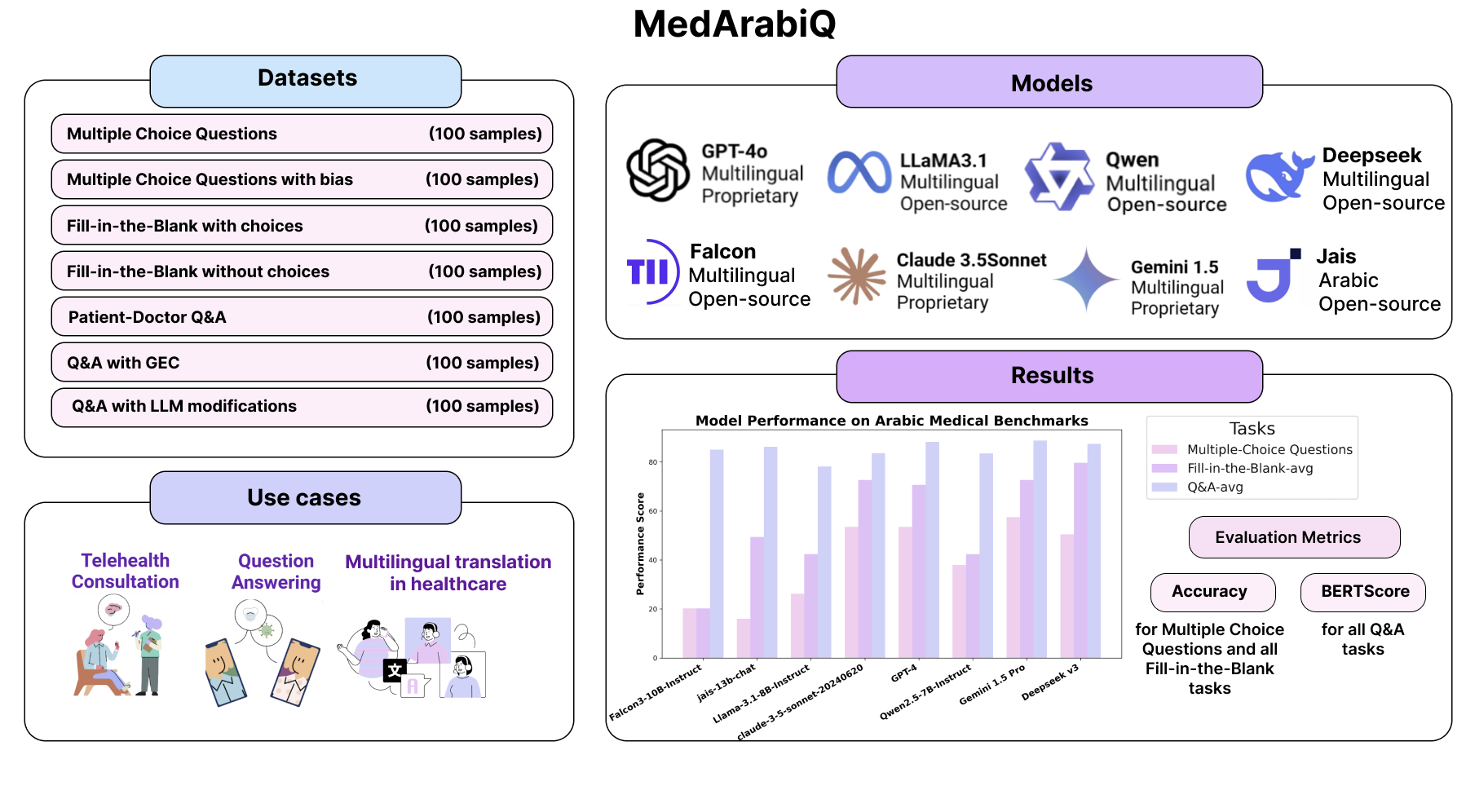}
    \caption{\textbf{Overview of MedArabiQ.} We construct seven new benchmark datasets and evaluate eight state-of-the-art LLMs on them.}
    \label{fig:overview}
\end{figure}

In this section, we describe the details of our methodological framework to construct the datasets and evaluate state-of-the-art LLMs. We start by explaining our chosen tasks, models and the experimental setup used to evaluate model performance. In Section \ref{sec:cohort}, we further explain our data collection and preprocessing methods. A general overview of MedArabiQ is provided in Figure~\ref{fig:overview}.
\color{black} 

\subsection{Tasks and Models}
In order to develop a reliable framework for evaluating LLMs in Arabic healthcare applications, we focused on telehealth consultation and question-answering as the main use cases. These involve not only medical reasoning ability but also natural patient-doctor dialogue. The LLM should mimic the role of a clinician as closely as possible, which includes possession of medical knowledge and the ability to utilize the knowledge, personalized to the needs and queries of patients. Personalization, however, should not lead to bias or prejudice in the LLM's response, based on the patient's profile. 

At the same time, it is important to note that benchmarks used to evaluate LLMs often face the challenge of data contamination. Many modern LLMs are trained on large web crawl datasets sourced from the internet, which can inadvertently include benchmark questions. This overlap inflates performance metrics, making such benchmarks unreliable for fair evaluation \citep{deng2024, dong2024}. Addressing this issue is critical to ensure that benchmark results accurately reflect the true capabilities of LLMs, free from biases introduced by prior exposure to evaluation data.

To mitigate these concerns, we categorize models into two primary groups: those with known training data and those with unknown training data. This initial distinction allows us to assess contamination risks more reliably for models whose training sources are documented, while acknowledging that for models without such transparency, contamination cannot be easily ruled out.

In addition to categorizing models based on the transparency of their training data, this study further classifies models according to their linguistic coverage. Specifically, we distinguish between models that are broadly multilingual and those that are primarily Arabic-centric multilingual. Although many LLMs include training data from numerous languages, exposure to a language does not inherently confer proficiency \citep{bender2021}. GPT-4, for example, is known to have been trained on a variety of languages \citep{achiam2023gpt}. By assessing multilingual and Arabic-centric multilingual models separately, we explore how language coverage and specialization influence performance, particularly in the domain of Arabic medical tasks. Our model grouping is shown in Table \ref{table:models} of Appendix \ref{appendix-b}.

This grouping reflects important distinctions in model transparency and linguistic specialization, but we acknowledge the considerable heterogeneity within each group regarding architecture, scale, training data, and performance. To better account for this, we provide additional analysis in Figure~\ref{fig:comp_model} and Table~\ref{tab:model_comparison} of the appendix, comparing models by training compute (FLOPs) and dataset size.
As expected, larger training compute generally correlates with improved performance, particularly for closed models such as GPT-4 and Gemini 1.5 Pro. Training dataset size shows some correlation within open models, though trends are less consistent. These analyses support the rationale for our grouping by offering further context for interpreting performance differences amid diverse model characteristics.

\subsection{Experimental Setup}
Our framework evaluates LLMs on Arabic medical tasks through seven datasets derived from two sources: (1) private structured medical exams/notes and (2) public patient-doctor dialogues adopted from AraMed \citep{alasmari2024}, as will be shown in further detail in the following sections.
\color{black}

\textbf{Evaluation.} The evaluation of LLMs in Arabic healthcare applications requires a comprehensive framework that balances technical performance with real-world applicability \citep{sallam2024}. Our framework assesses medical reasoning, decision-making, and dialogue-based question-answering \citep{kanithi2024, guo2023}. Multiple-choice questions (MCQ) were evaluated on accuracy of the answer, while open-ended questions, such as those in the fill-in-the-blank and patient-doctor Q\&A, were assessed using BERTScore to measure semantic alignment \citep{bert-score}. We perform zero-shot prompting for all models and tasks. 

Rather than applying a uniform temperature across models and tasks, we tuned temperature settings based on empirical testing and guidance from prior work \citep{du2025}, which recommends task- and model-aware configuration to improve generation quality without introducing evaluation bias. For classification tasks such as MCQ and fill-in-the-blank, we used a low temperature (0.0 for open models, 0.2 for closed models) to promote deterministic, stable outputs. For generative tasks like patient-doctor Q\&A, slightly higher temperatures improved coherence and naturalness without degrading factuality. Accordingly, we used 0.4 for open models (e.g., LLaMA 3.1, Qwen 2.5) and 0.2 for closed models (e.g., GPT-4, Gemini 1.5 Pro). We further extended our experiments with a temperature ablation study across representative models in Appendix \ref{app:tempstudy}. For semantic evaluation, we use the XLM-RoBERTa-Large model in BERTScore computation due to its multilingual training, which includes Arabic, making it more appropriate than English-centric or monolingual Arabic models.

\textbf{Instruction-tuned Models.} We employed instruction-tuned versions of the models described earlier due to their superior ability to interpret and execute task-specific instructions. In contrast, the base versions demonstrated significant limitations in following prompts, even with extensive prompt engineering. This aligns with the literature \citep{chung2022scaling, zhang2023instruction}, which shows that instruction fine-tuning significantly improves model performance and prompt adherence across various tasks and model sizes.

\textbf{Instruction Prompts.} Prompt engineering is crucial for evaluating LLMs. In our prompt engineering experiments, we tested both English and Arabic prompts and found that English prompts were generally more effective. However, for Q\&A benchmark tasks, Arabic prompts performed better in open-access models. Based on these findings, we used English prompts for all tasks except for the Q\&A datasets.  The same prompt was used for each model, though it was customized to each category of tasks as shown in Table~\ref{table:prompts} of the appendix. This was crucial to ensure the prompt reflected realistic scenarios as closely as possible, tailored to different use cases.

\textbf{Answer Processing.} For MCQ and fill-in-the-blank with choices tasks, models generate both the index of the predicted correct option and the full textual answer. However, some models, particularly open-access ones, exhibit spelling inconsistencies. To ensure accuracy, we evaluated only the first character generated after the phrase ``The correct letter is:'', comparing it against valid options. If the character does not match a valid choice, then the response is deemed invalid. For open-ended benchmarks, where answers are not constrained to multiple-choice formats, we evaluate the full model-generated response.

\subsection{Bias Assessment and Mitigation}
To ensure that LLMs can be effectively deployed in healthcare and provide meaningful contributions to the field, it is critical to address their potential to replicate human biases. Completely eliminating bias from LLMs is nearly impossible, as the datasets used to train these models are inherently shaped by human judgment, which is itself susceptible to bias. Recognizing this challenge, we developed an evaluation framework to systematically assess model resistance to biases injected into MCQ, measure susceptibility, and evaluate mitigation strategies. The framework draws on methodologies outlined in recent work~\citep{schmidgall2024addressing}, while adapting key elements to the context of Arabic medical datasets and healthcare applications. These adaptations include culturally relevant bias categories, prompts designed for clinical scenarios, and additional evaluation metrics to align with real-world needs.

We created a structured framework to systematically assess  language-model resistance to cognitive biases:

\begin{enumerate}
    \item \textbf{Baseline Testing:} Models are evaluated using the original, unbiased dataset to establish baseline performance metrics.
    
    \item \textbf{Bias Testing:} Models are tested with biased variations of prompts. We compute the change in accuracy to assess impact of bias (see Table~\ref{tab:bias_examples} of the appendix for bias injection examples).
    
    \item \textbf{Evaluation of Bias Mitigation:} Mitigation techniques are tested to determine their impact on accuracy. These include:
    \begin{itemize}
        \item \textbf{Bias Education:} Adding warnings to prompts emphasizing evidence-based reasoning (e.g., ``Evaluate each patient uniquely, without relying on trends or recent cases'').
        \item \textbf{One-Shot Demonstration:} A single negative example is provided to illustrate incorrect reasoning caused by bias.
        \item \textbf{Few-Shot Demonstration:} Both negative and positive examples are presented to show correct and incorrect handling of bias.
    \end{itemize}
    See Table~\ref{table:mitigation_example} in the appendix for an illustration of these mitigation strategies in use.
\end{enumerate}

This framework provides a systematic and reproducible approach to assess and address cognitive biases in LLMs, ensuring their deployment in healthcare contexts is both effective and ethically sound.
\newline
\section{Dataset Construction}
\label{sec:cohort}
As previously stated, our framework for evaluating LLMs in Arabic healthcare applications integrates clinically validated knowledge and linguistically diverse sources to construct a benchmark that mirrors real-world patient needs and clinical reasoning challenges. 
\color{black}
We do so by deriving our datasets from two primary sources: {past exams and notes from Arabic medical schools}, and the {AraMed Dataset} \citep{alasmari2024}.

\subsection{Data Selection} 
In our data selection process, we specifically selected data sources that were unlikely to have been included in prior training datasets. 
\subsubsection{Multiple-Choice Questions}
\label{mcq}
To evaluate the models' medical understanding, we curated a standard dataset with question-answer pairs, covering foundational and advanced medical topics, such as physiology, anatomy, and neurosurgery. From our curated dataset, we selected a random set of 100 multiple-choice questions on an ad-hoc basis by prioritizing: (i) broad coverage of specialties, (ii) varying difficulty levels, and (iii) clarity of question and answer. We then digitized and extracted them into CSV files and performed manual verification. The average question length is 24 words. 

\subsubsection{Multiple-Choice Questions with Bias}
Following recent work \citep{schmidgall2024addressing}, we injected bias in the multiple-choice questions dataset to evaluate how LLMs handle ethical or culturally sensitive scenarios. In particular, we utilized a set of well-defined bias categories \citep{schmidgall2024addressing}, including (i) confirmation bias, (ii) recency bias, (iii) frequency bias, (iv) cultural bias, (v) false-consensus bias, (vi) status quo bias, and (vii) self-diagnosis bias. By manually injecting the bias, we ensured relevance to the unique linguistic and clinical challenges of Arabic healthcare contexts. This resulted in a dataset consisting of 100 samples.

\subsubsection{Fill-in-the-Blank with Choices}
\label{fitb-choices}
To assess knowledge recall and in-context learning, we manually constructed fill-in-the-blank questions, each accompanied by a set of predefined answer choices. The model was required to select the most appropriate answer from the given options. This approach evaluates the model's ability to recognize correct answers within a constrained set, reducing the reliance on generative capabilities. The resulting dataset consists of 100 samples.

\subsubsection{Fill-in-the-Blank without Choices}  
In this setting, the fill-in-the-blank questions were presented without predefined answer choices, requiring the model to generate responses independently. This evaluation measures the model's ability to recall and generate accurate medical knowledge without additional information, emphasizing its reasoning and language generation capabilities. The dataset for this task also comprises 100 samples.

\subsubsection{Patient-doctor Q\&A}
AraMed is an Arabic medical corpus for question-answering that was originally sourced from AlTibbi, an online medical patient-doctor discussion forum \citep{alasmari2024}. The original dataset consists of 270,000 question-answer pairs, of which 400 were made publicly available. We meticulously selected 100 samples covering the entire range of specialties present originally in AraMed, ensuring quality and avoiding redundancy.

\subsubsection{Q\&A with Grammatical Error Correction (GEC)}
 Since the patient-doctor Q\&A dataset uses dialectal Arabic, we constructed an additional version with enhanced linguistic quality and consistency. We specifically applied a Grammatical Error Correction (GEC) pipeline tailored for Arabic healthcare texts. Given the morphological complexity and syntactic richness of Arabic, this preprocessing step was essential. First, we used cameltools, an open-source Arabic NLP library, to morphologically disambiguate words and dediacritize text, ensuring uniformity \citep{obeid2020}. This step was crucial for handling Arabic’s inflectional patterns and preparing the dataset for grammatical correction. Then, we employed an existing fine-tuned BERT-based Arabic Grammatical Error Detection (GED) model~\citep{alhafni-etal-2023-advancements} to detect agreement errors, incorrect word order, and missing inflections. Each token was tagged with a grammatical label to facilitate structured corrections. Next, we used the GEC model from the same pipleine, built on mBART, to automatically correct detected errors while preserving semantic integrity. The model was fine-tuned on the QALB-2015, QALB-2014, and ZAEBUC corpora to enhance its performance in grammatical corrections \citep{mohit-etal-2014-first, rozovskaya-etal-2015-second, habash-palfreyman-2022-zaebuc}.
 
\subsubsection{Q\&A with LLM Modifications}
\label{qa-llm}
Additionally, to mitigate potential model memorization, since some models could have been trained using scraped data of Altibbi, we modified the dataset using an LLM for a more rigorous assessment of reasoning and adaptability \citep{dong2024}. In particular, we used GPT-4o to paraphrase our original questions using the prompt, ``You are a helpful assistant that paraphrases text while keeping its meaning intact.'' This approach ensured that the core medical concepts remained unchanged while reducing the likelihood of models relying on memorized content \citep{zhu2025}.

In summary, we constructed seven new datasets using two primary sources, AraMed and past medical exams, via extensive manual verification to build the MedArabiQ benchmark, all of which are summarized in Table \ref{table:datasets} in Appendix \ref{appendix-f}.

\subsection{Data Extraction} 
In the collection of the MCQ dataset in Section \ref{mcq} and the fill-in-the-blank questions dataset in Section \ref{fitb-choices}, we started off by collecting paper-based past exams and lecture notes sourced from a large repository of academic materials hosted on student-led social platforms of regional medical schools. No personally identifiable information or real patient data was included, and thus anonymization was not necessary as our data collection complied with privacy and ethical guidelines. These exams and answers were not readily available in structured digital formats, necessitating a rigorous manual process to ensure clarity and correctness. Given that Arabic medical education is not widely digitized, these exams are not publicly accessible in structured formats. Even if some individual questions exist online, the extensive effort required to compile, format, and structure them into a benchmark dataset significantly reduces the likelihood of contamination. Questions were selected to reflect increasing complexity across different academic years, ensuring that model performance could be assessed at varying levels of medical expertise.

As for the patient-doctor Q\&A dataset, our selection process prioritized questions with well-formed queries and meaningful answers, avoiding instances where responses were overly generic (e.g., ``Consult a doctor'' or ``See a specialist''). However, we retained some examples of such cases to reflect real-world user behavior, as patients often seek medical advice even for questions that are better suited for in-person consultations.

Additionally, we maintained proportionality in representation to mirror the dataset's original structure, particularly for categories like reproductive and sexual health. These categories comprised a significant portion of the dataset and are often underrepresented in Arabic-language medical research despite their importance. Given the sensitivity and cultural taboos associated with these topics, we ensured their inclusion to provide a more realistic and balanced evaluation of the model's ability to handle diverse medical inquiries. This approach ensures the dataset reflects the varied nature of medical concerns while maintaining its relevance to real-world healthcare scenarios. We selected 100 questions, allocating an equal proportion of samples to each specialty: cardiology, obstetrics and gynecology, surgery, pediatrics, neurology, oncology, endocrinology, dentistry, otholarhyntology, public health, dermatology, primary care, pulmonology, and psychology.

We also incorporated information about the patient, specifically age and gender, when known, into the query. When this information was unknown or erroneous (i.e., the age was an illogical number), it was excluded. Some questions came from an acquaintance of the patient, so no information was known about the patient. The information was usually prepended to the question or inserted after the greeting, in the format of ``I am a [man/woman] and I am [x] years old''. The inclusion of this information was useful in incorporating personalization into the case scenarios, making our benchmark dataset better suited to evaluate LLM performance in real-world medical scenarios.


\color{black}

\section{Results} 
We report all results of the experiments performed on six of our benchmarks in Table~\ref{table:results}. Our results show that no single model outperforms all others across all benchmarks. For closed tasks (MCQ and fill-in-the-blank with and without choices), closed models perform best, as expected. Gemini 1.5 Pro achieves the highest accuracy in two out of six benchmark tasks, while Deepseek leads on the fill-in-the-blank with choices task with an accuracy of 79.7. For closed tasks, Gemini and Deepseek perform best with accuracy scores of 57.5 and 79.7, respectively. In open-ended tasks, namely patient-doctor Q\&A, Q\&A with GEC, and Q\&A with LLM modifications, Jais performs best, achieving scores of 85.7, 85.6, and 85.5 respectively. However, models do not consistently excel across both task types, as Jais performs poorly on closed tasks. Among open-access models, Jais is the best-performing, followed by LLaMA, which achieves the highest score on Q\&A with LLM Modifications.

\begin{table*}[t!]
\setlength{\tabcolsep}{2pt}
\floatconts
{tab:benchmark_results}
{\caption{\textbf{Summary of performance results across all benchmark datasets.} We present the results in terms of accuracy and BERTScore, depending on task, and show best results in bold per row. Due to space constraints, abbreviated model names are used in the table; full names are as follows: Falcon = \texttt{Falcon3-10B-Instruct}, Jais = \texttt{jais-13b-chat}, LLaMA = \texttt{LLaMA-3.1-8B-Instruct}, Claude = \texttt{claude-3-5-sonnet-20240620}, GPT-4 = \texttt{gpt-4-0613}, Qwen = \texttt{Qwen2.5-7B-Instruct}, Gemini = \texttt{Gemini 1.5 Pro}, Deepseek = \texttt{Deepseek v3}.}}
{\resizebox{\textwidth}{!}{ 
\begin{tabular}{llcccccccc}
\toprule
\textbf{Benchmark Datasets} & \textbf{Metrics}  & \textbf{Falcon} & \textbf{Jais} & \textbf{LLaMA} & \textbf{Claude} & \textbf{GPT-4} & \textbf{Qwen} & \textbf{Gemini} & \textbf{Deepseek}\\
\midrule
\multirow{1}{*}{\makecell{Multiple-Choice Questions}} & Accuracy & 20.2 & 16.0 & 26.2 & 53.5 & 53.5 & 38.0 & \textbf{57.5} & 50.5\\ 
\midrule
\multirow{1}{*}{\makecell{Fill-in-the-Blank with choices}} & Accuracy   & 20.2 & 49.4 & 42.4 & 72.7 & 70.7 & 42.40 & 72.7 & \textbf{79.7}\\
\midrule
\multirow{1}{*}{\makecell{Fill-in-the-Blank without choices}} & BERTScore  & 85.1 & 86.2 & 78.2 & 83.6 & 88.2 & 83.5 & \textbf{88.8} & 87.4\\
\midrule
\multirow{1}{*}{\makecell{Patient-Doctor Q\&A}} & BERTScore      & 84.9 & \textbf{85.7} & 81.2 & 81.1 & 84.5 & 85.2 & 82.5 & 82.2 \\
\midrule
\multirow{1}{*}{\makecell{Q\&A with 
LLM modifications}} & BERTScore    & 84.8 & \textbf{85.6} & 85.5 & 83.7 & 84.5 & 85.2 & 82.5 & 82.3\\ 
\midrule
\multirow{1}{*}{\makecell{Q\&A with Grammatical Error Correction}} & BERTScore   & 84.4 & \textbf{85.5} & 84.9 & 83.6 & 84.2 & 84.9 & 82.3 & 82.0\\

\bottomrule
\end{tabular}
}}
\label{table:results}
\end{table*}

\begin{figure*}[t!]
    \centering
    \includegraphics[width=0.99\linewidth]{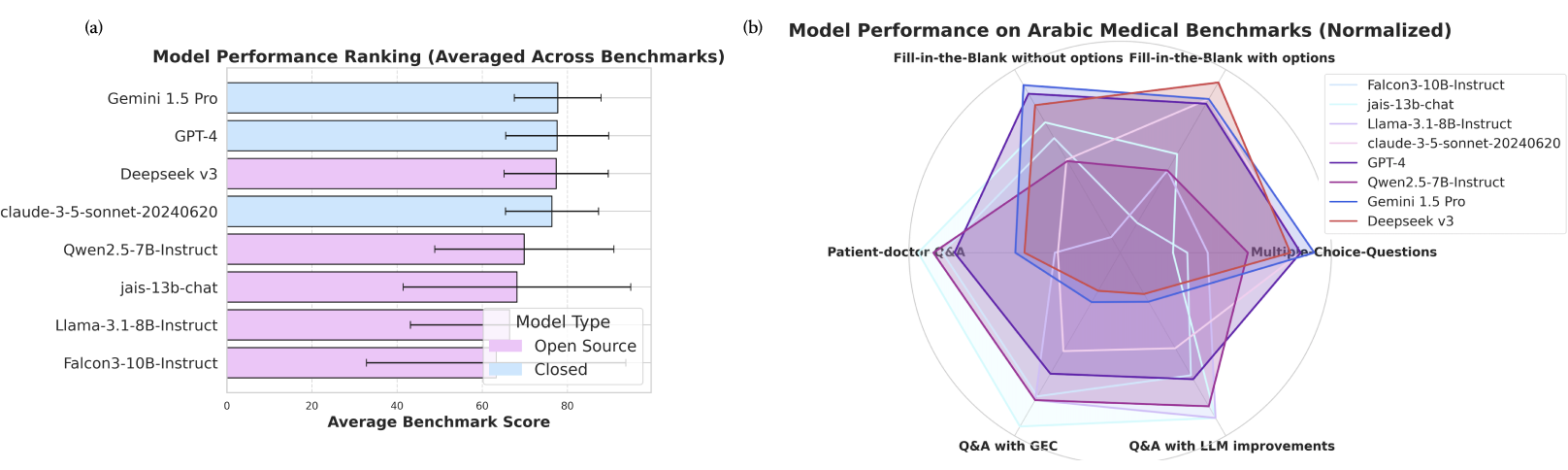}
    \caption{\textbf{Overall results of SOTA models on the MedArabiQ benchmark.} (a) Average performance across  benchmark datasets. (b) Breakdown of model performance across  benchmark tasks.}
    \label{fig:results_overall}
\end{figure*}

\begin{figure*}[t!]
    \centering  \vspace{-5mm}
    \includegraphics[width=0.99\textwidth]{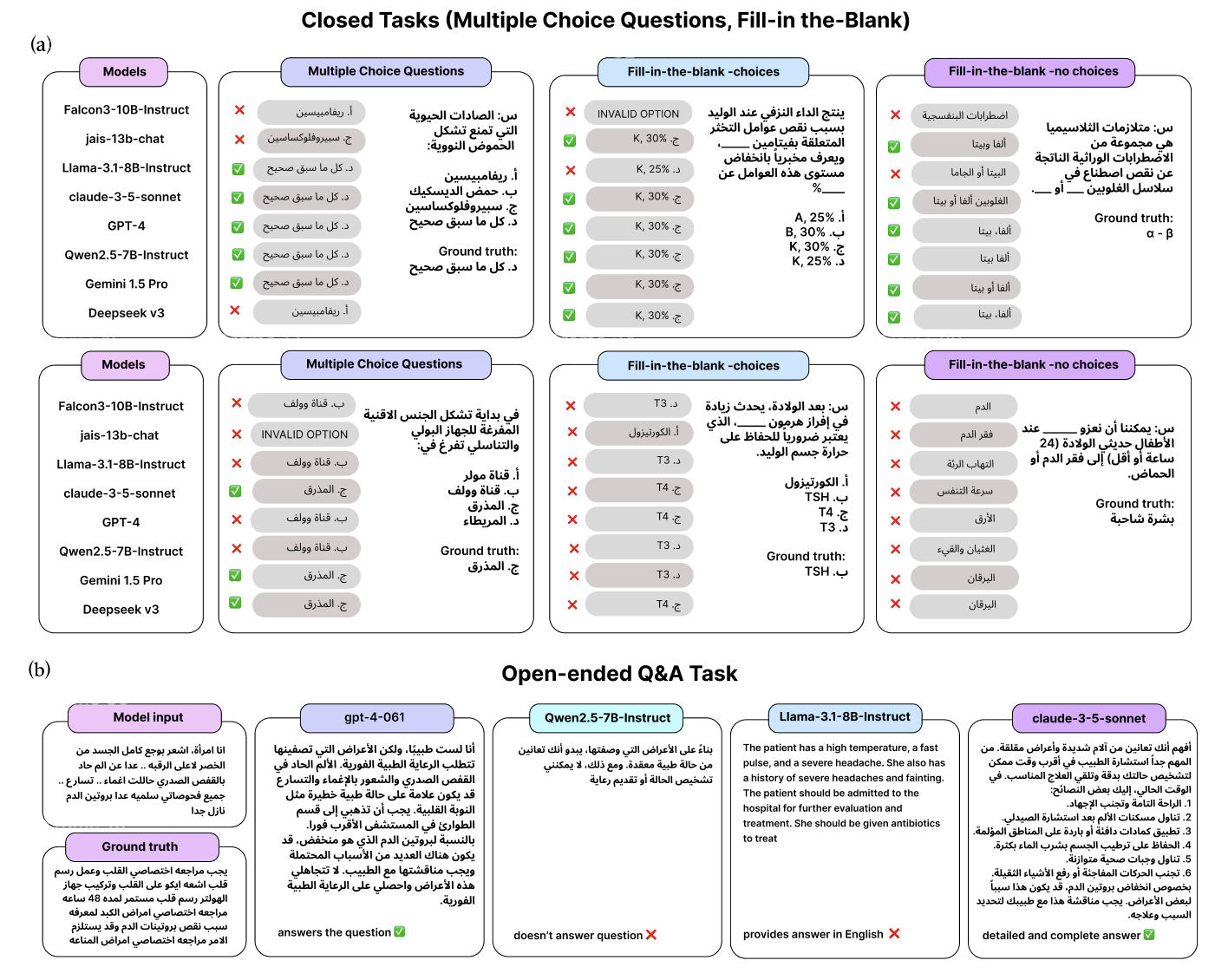} 
    \vspace{-5mm}
    \caption{\textbf{Performance Samples from Closed and Open-ended Benchmarks.} We report model outputs for samples from (a) MCQ and Fill-in-the-Blank, and (b) patient-doctor Q\&A tasks, illustrating differences in accuracy, response validity, and language consistency across proprietary high-resource and open-access models. An English version of this figure is provided in Figure~\ref{fig:samples_english} of the appendix.}
    \label{fig:samples-new}
\end{figure*}

\begin{figure*}[ht!]
    \centering
    \includegraphics[width=0.8\textwidth]{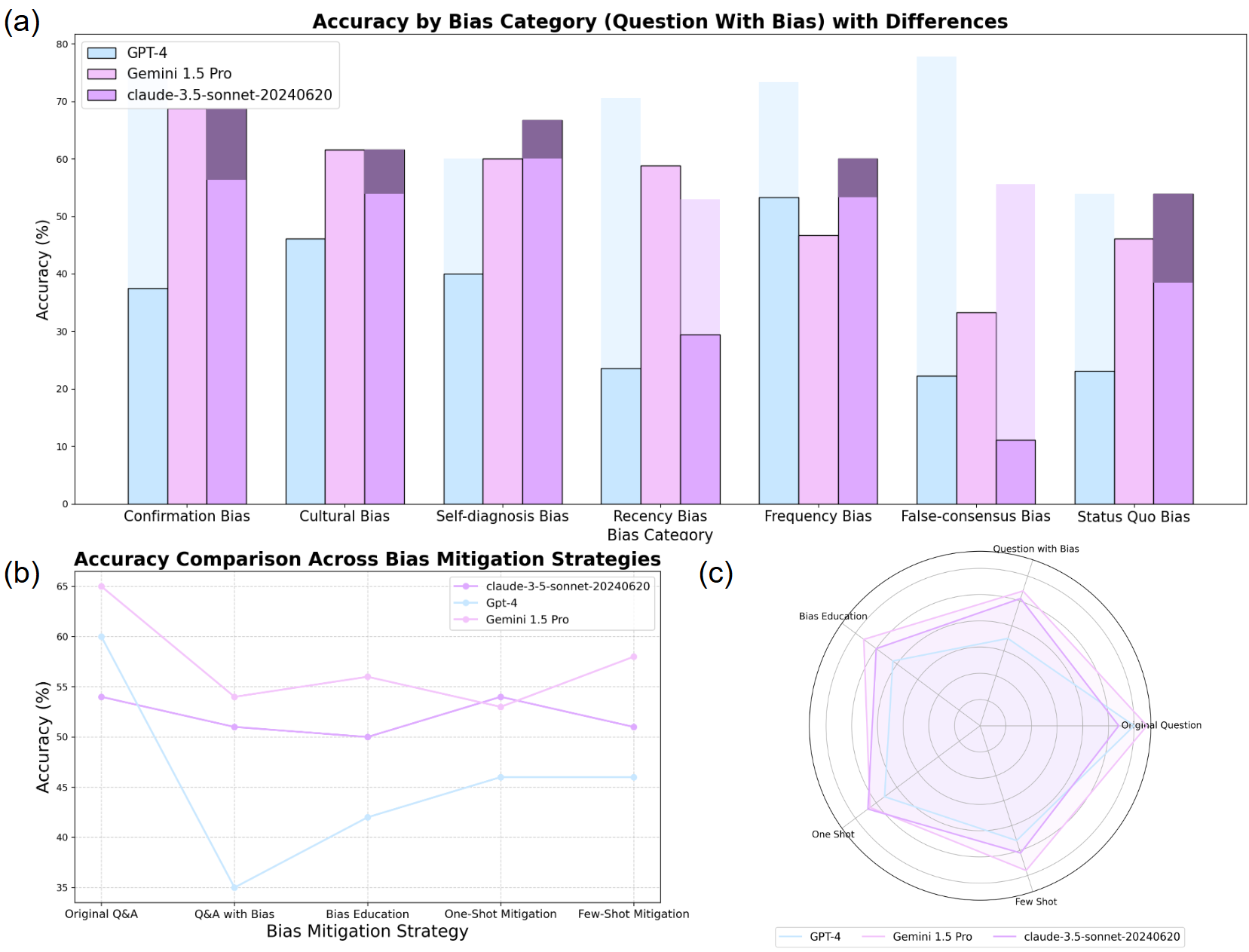} 
    \caption{\textbf{Overview of Bias Mitigation Strategy Results for top-performing models.}
    (a) Comparison of accuracy between original questions and questions with bias across different bias categories. The bars represent accuracy percentages, with
    lighter shades representing a decrease, and darker shades an increase, in accuracy after bias incorporation.
    (b) Line plot comparing accuracy scores across various mitigation strategies.
    (c) Radar plot illustrating model performance across different bias mitigation strategies, highlighting relative strengths and weaknesses of each model in handling bias-related challenges.
    }
    \label{fig:bias}
\end{figure*}

Figure~\ref{fig:results_overall} (a) compares the performance of open-access and proprietary high-resource models averaged across all benchmark tasks. Among all models, \textbf{Gemini 1.5 Pro} achieves the \textbf{highest average benchmark score}, followed closely by GPT-4, Deepseek v3 and Claude 3.5, reinforcing the dominance of proprietary high-resource models in NLP tasks. The open-access models, including Llama 3.1, Qwen 2.5, and Falcon, demonstrate competitive but more variable performance, with Qwen 2.5 emerging as the strongest among them. The error bars indicate that while closed models exhibit greater stability across tasks, open-access models show higher performance variability, likely due to their reliance on general-purpose pretraining rather than domain-specific fine-tuning. These findings highlight the current dominance of proprietary high-resource models while underscoring the potential of open-access models with further fine-tuning and domain adaptation.

Figure~\ref{fig:results_overall} (b) provides a task-specific performance breakdown, revealing significant performance variation across MCQ, fill-in-the-blank, and generative Q\&A tasks. Closed models consistently achieve high scores across all benchmarks, particularly excelling in Q\&A tasks, including those with GEC and LLM-modified questions. This suggests a strong capability for handling complex medical inquiries, a critical requirement in real-world applications. However, the fill-in-the-blank and multiple-choice tasks exhibit more performance divergence, with open-access models like Qwen 2.5 and Llama 3.1 lagging behind the leading closed based models. Notably, Jais performs better in certain benchmarks but struggles with structured tasks, suggesting limitations in factual consistency and retrieval-based reasoning. The observed discrepancies emphasize the need for fine-tuning on structured medical datasets, particularly for open-access models, to improve accuracy in knowledge-intensive tasks. Figure~\ref{fig:samples-new} shows examples of model responses for the closed and open-ended tasks.

\begin{table*}[htbp]
\floatconts
{tab:bias_evaluation_results}
{\caption{\textbf{Comparison of performance by accuracy score without bias, with bias, and with bias mitigation across the three chosen models.}}}
{\resizebox{\textwidth}{!}{
\begin{tabular}{llccccc}
\toprule
\textbf{Model} & \textbf{Q\&A} & \textbf{Q\&A with Bias} & \textbf{Bias Education} & \textbf{One-Shot Mitigation} & \textbf{Few-Shot Mitigation} \\
\midrule

Claude 3.5 Sonnet  & 53.1 & 52.0 & 51.0 & \textbf{55.1} & 52.0\\
GPT-4              & \textbf{66.3} & 35.7 & 42.9 & 46.9   & 46.9   \\
Gemini 1.5 Pro     & 55.1 & \textbf{55.1} & \textbf{57.1} & 54.1 & \textbf{59.2} \\

\bottomrule
\end{tabular}
}}
\label{table:bias}
\end{table*}

Table \ref{table:bias} and Figure~\ref{fig:bias} highlight the impact of bias and different mitigation strategies on model performance. In Figure~\ref{fig:bias} (a), we compare the accuracy of GPT-4, Gemini 1.5 Pro, and Claude 3.5 Sonnet-20240620 on original questions versus questions injected with bias across various bias categories. The results reveal that, generally, all models experience a decline in accuracy when bias is introduced, with the extent of decrease varying by bias type. Across all bias catgories, Gemini 1.5 Pro demonstrates high resilience to bias, while GPT-4 shows more significant performance drops in False-Consensus Bias and Recency Bias. Additionally, Figure~\ref{fig:bias} (b) illustrates the effectiveness of different bias mitigation strategies. Notably, all models exhibit improved accuracy with strategies like Few-Shot prompting compared to biased questions without mitigation. Gemini 1.5 Pro outperforms the other models with both Bias Education and Few-Shot Mitigation, reinforcing its robustness in handling biased inputs.

Figure~\ref{fig:bias} (c) presents a radar plot summarizing model performance across the employed bias mitigation strategies. The plot shows Gemini's consistent performance across all strategies, while GPT-4 and Claude 3.5 Sonnet display larger variability across strategies such as Bias Education and One-Shot prompting. These results further highlight the importance of bias mitigation strategies in enhancing model reliability. Additional information on the performance of models by bias category and medical specialty across different bias mitigation strategies can be found in Figures~\ref{fig:bias accuracy1} and~\ref{fig:bias accuracy2} of the appendix, respectively.

\section{Discussion} 
Our findings are consistent with prior research on medical LLM benchmarking, where proprietary high-resource models outperform open-access models in structured tasks but perform comparably in generative ones. \cite{chen2025benchmarkinglargelanguagemodels} demonstrated that proprietary high-resource models like GPT-4 and Med-PaLM 2 excel in multiple-choice and factual retrieval tasks, largely due to stronger pretraining on structured datasets and domain-specific knowledge integration. Similarly, \cite{Alonso_2024} found that proprietary high-resource models generally achieve higher accuracy in medical question-answering tasks across multiple languages, reinforcing the idea that closed models have better knowledge grounding and factual consistency. Our results support these trends, with Gemini 1.5 Pro and Claude 3.5 Sonnet leading in multiple-choice and fill-in-the-blank tasks, suggesting that API-based models are better suited for clinical decision support and structured question-answering tasks.

Jais underperforms even compared to a random baseline in structured tasks due to severe hallucinations, generating unreliable responses, while Falcon struggles to generalize due to limited exposure to high-quality medical datasets\citep{penedo2023}. Figure~\ref{fig:samples-new} further illustrates these issues, particularly in the fill-in-the-blank task, where Jais frequently selects invalid options or provides factually inconsistent answers. These results highlight the limitations of open-access models in structured medical applications, where their general-purpose pretraining leads to factual inconsistencies, necessitating domain-specific fine-tuning. Additionally, the relatively lower performance of the Arabic-trained Jais relative to other open-access multilingual models such as Qwen suggests that task specificity may be more critical than language specificity, for Arabic medical tasks, and warrants further exploration in the future.

Figure~\ref{fig:bias} further illustrates the variability in model performance when exposed to biased content and the effectiveness of different bias mitigation strategies for the top 3 scoring models (GPT-4, Gemini 1.5 Pro, and Claude 3.5 Sonnet-20240620). All models experience an accuracy drop when bias is introduced, with GPT-4 Sonnet showing the most pronounced decline across multiple bias categories. Notably, bias mitigation techniques such as One-Shot and Few-Shot prompting have shown significant improvement in model performance, especially for Gemini 1.5 Pro, which shows the greatest resilience against bias. 
Yet, no single strategy consistently mitigates bias across all models and categories, underscoring the complexity of bias in medical NLP and the need for further research.

For generative Q\&A tasks, however, traditional automatic evaluation metrics such as BERTScore do not fully capture actual model performance. Although GPT-4 and Claude 3.5 produce responses that are highly relevant and contextually accurate as shown in Figure \ref{fig:samples-new}, their longer responses result in lower BERTScores compared to ground truth references. This issue has been previously noted by \cite{liu-etal-2025-interactive} who found that metrics like ROUGE and BERTScore struggle to effectively assess medical LLMs due to the inherent complexity of medical diagnoses, where multiple treatment options may exist for a single patient. Our findings reinforce this limitation, as models with lower BERTScores sometimes generate high-quality, informative answers that are penalized for verbosity rather than inaccuracy. 

To address this limitation, we conducted an additional analysis using GPT-4 as an LLM-based judge, following prior work \citep{thakur2025}. Each model's response was rated on a scale of 1 to 5 across four dimensions: similarity to ground truth, relevance/helpfulness, factuality, and safety. This evaluation revealed discrepancies with BERTScore rankings. For example, while Jais had the highest BERTScore (85.7), it received an average GPT score of only 3.2, with models like Deepseek (4.2), Gemini (4.1), and GPT-4 (3.9) outperforming it---particularly in factuality and relevance. Falcon similarly scored well under BERTScore but received the lowest human-aligned rating (1.1), indicating the metric’s failure to capture hallucinations. We also examined potential self-enhancement bias and found none, as GPT-4 rated other models more favorably than itself. These findings, detailed in Appendix~\ref{sec:llm-as-judge}, highlight the limitations of surface-level overlap metrics and underscore the value of more nuanced evaluation techniques such as expert assessments or task-oriented dialogue evaluation for real-world deployment in medical settings.

\paragraph{Ethical Considerations}
Given the sensitive nature of healthcare applications, ethical considerations are paramount when developing and deploying LLMs. Our benchmark, MedArabiQ, was carefully constructed from publicly available and anonymized educational resources, ensuring compliance with data privacy standards by excluding personally identifiable patient information. Furthermore, we recognize and discuss several ethical implications associated with clinical deployment of LLMs, including potential risks such as misinformation, unintended amplification of biases (cultural, confirmation, or status quo biases), and limited model interpretability. Addressing these concerns, we emphasize the importance of comprehensive validation through hybrid frameworks combining automated evaluation metrics with clinician expert reviews. We advocate for continuous monitoring, clinician oversight, and clearly defined operational guidelines to mitigate potential harm and biases. By explicitly detailing these ethical considerations, we aim to promote responsible and transparent adoption of LLMs, ultimately contributing to safer and more equitable healthcare solutions.

\paragraph{Limitations}
While our study provides a comprehensive evaluation of Arabic medical LLMs, there are areas that warrant further exploration. First, there is a possibility for contamination in the datasets. Although the past medical exams are not available in structured digital format and required an extensive effort for digitization and cleaning, we cannot completely rule out the possibility of contamination. That being said, model performance can be improved highlighting the importance of the benchmark datasets. As for the patient-doctor Q\&A data, we explicitly incorporated modifications to test for potential memorization considering that it is sourced from a publicly available dataset. To verify the validity of the dataset, preliminary results were obtained in collaboration with medical students to assess the Q\&A dataset for relevance, factuality, complexity, and clarity on a scale from 1 to 5. The average scores (Relevance: 4.99, Factuality: 4.97, Accuracy: 4.88, Clarity: 4.89) provide initial evidence of the dataset's reliability and utility. Extensive evaluation is required in future work as the scope and size of the benchmark is expanded. A systematic human evaluation, such as expert annotation, would further strengthen the dataset's credibility, and we plan to incorporate this in future releases.

Additionally, following standard practices in medical NLP, we rely on benchmark datasets rather than live clinical interactions, ensuring reproducibility and ethical compliance. Notably, our patient-doctor Q\&A dataset is sourced from AraMed \citep{alasmari2024}, which consists of real consultations conducted on the Altibbi telemedicine platform, ensuring real-world relevance. While live clinical testing could offer additional insights, it presents substantial privacy and regulatory challenges, particularly in the Arab region, where GDPR and HIPAA regulations impose strict limitations on patient data sharing \citep{theodos2020}. Additionally, many healthcare facilities in the region still rely on paper-based records \citep{aljawarneh2024}, making large-scale real-time data collection complex. Future work could explore privacy-preserving strategies for integrating real-world clinical assessments in a secure and ethical manner.

Another limitation of our study is the need for more robust bias mitigation techniques. While we evaluate bias susceptibility across multiple tasks, our findings reinforce that existing strategies do not fully eliminate bias, particularly in sensitive clinical decision-making contexts \citep{alyafeai2024, omar2024}. Future research should focus on developing domain-specific bias mitigation approaches that address linguistic and cultural factors unique to Arabic medical NLP.

\paragraph{Future Work}
In this study, we focus on evaluating zero-shot performance, providing an unbiased assessment of how well LLMs perform on Arabic medical tasks without prior adaptation \citep{kojima2022}. While this establishes a strong baseline, future work could explore fine-tuning on Arabic medical datasets to further enhance domain-specific understanding. Fine-tuning presents an opportunity to improve model accuracy and contextual relevance, though it also requires careful consideration of data availability, computational costs, and generalization risks.

Furthermore, our benchmarks are in MSA, following standardization practices to ensure consistency. However, this does not account for dialectal variations, which can be problematic in real-world patient-doctor interactions. Future work could explore incorporating dialectal data to enhance model adaptability across diverse Arabic-speaking healthcare contexts. Additionally, while this study focuses on text-based benchmarks as a unimodal foundation, expanding Arabic medical NLP benchmarks to support multimodal inputs---such as medical images and lab results---would be a valuable direction for future research. 

In the future, we aim to actively expand MedArabiQ by increasing both the breadth and depth of medical specialties covered. Specifically, we plan to broaden the benchmark to include more specialized clinical domains such as mental health, infectious diseases, and chronic illnesses. This expansion will involve close collaboration with expert physicians and medical educators who will help ensure the questions reflect current clinical standards and practices. Furthermore, we intend to classify questions into specific clinical reasoning types (e.g., diagnostic reasoning, treatment planning, patient counseling), systematically rate their complexity, and conduct rigorous inter-rater reliability assessments. This structured expansion significantly enhances the benchmark's comprehensiveness, reliability, and utility for fine-tuning Arabic-focused and multilingual LLMs, ultimately improving their clinical applicability and supporting equitable healthcare outcomes for Arabic-speaking populations.

\paragraph{}Overall, in this work, we introduced the first structured benchmark for evaluating LLMs in Arabic healthcare, addressing a critical gap in Arabic medical NLP. Our benchmark consists of 700 diverse clinical samples, covering both structured medical knowledge assessments and real-world patient-doctor interactions. Beyond Arabic healthcare, our benchmark lays the foundation for developing benchmarks in other medically underserved languages, contributing to the global refinement of medical AI applications.

Our evaluation exposes critical limitations in current LLMs, including factual hallucinations in open-ended tasks and vulnerability to biases in clinical decision-making, reinforcing the need for robust bias mitigation strategies. Future work should explore fine-tuning LLMs on Arabic medical datasets, expanding benchmarks to capture dialectal variations, and developing effective bias mitigation strategies tailored to Arabic medical contexts. By releasing our benchmarks, we aim to foster further research in Arabic medical NLP, providing a foundation for trustworthy, unbiased, and effective AI-driven healthcare solutions.

\section{Acknowledgments}
This work was supported by the NYUAD Center for Artificial Intelligence and Robotics, funded by Tamkeen under the NYUAD Research Institute Award CG010, the Center for Cyber Security (CCS), funded by Tamkeen under the NYUAD Research Institute Award G1104, and the Meem Foundation. The research was carried out on the High Performance Computing resources at New York University Abu Dhabi (Jubail).

\bibliography{main}

\newpage
\appendix

\section{Related Work}
\label{appendix-a}
 \setcounter{table}{0}
\renewcommand{\thetable}{A\arabic{table}}

 \setcounter{figure}{0}
\renewcommand{\thefigure}{A\arabic{figure}}

Several Arabic medical datasets exists, but they each fall short in some way and most are focused on the task of medical question-answering. In Table \ref{table:dataset_comparison}, we summarize these datasets, their limitations, and how we overcome their limitations with our dataset. To summarize, our contribution is four-fold:
\begin{enumerate}
    \item Introducing a dataset consisting almost entirely of novel proprietary data that has not been seen by language models before, carrying a much lower risk of contamination.
    \item Extending data from traditional question-answering to more varied use cases through modalities like multiple-choice and fill-in-the-blank, testing medical knowledge and reasoning.
    \item Ensuring the quality of data points by manual human curation, inspection, and/or review.
    \item Capturing the spirit of real-world patient-doctor interactions using data sourced from the online medical platform Altibbi.
\end{enumerate}

\begin{table*}[h]
    \centering
    \caption{\textbf{Comparison of Various Arabic Medical Question-Answering Datasets}}
    {\renewcommand{\arraystretch}{1.3} 
    \resizebox{0.95\textwidth}{!}{ 
    \begin{tabular}{p{3cm} p{4cm} p{5cm} p{6cm} p{6cm}}
        \toprule
        \textbf{{Dataset Name}} & \textbf{{Authors}} & \textbf{{Description}} & \textbf{{Limitations}} & \textbf{How our paper addresses its limitations} \\
        \midrule
        \textbf{{CQA-MD}} & {\cite{nakov2019semeval2016task3community}} & {A corpus of over 45,000 question-answer pairs sourced from 3 online Arabic medical forums.} & {Though it is comprehensive, more than half of the question-answer pairs were noted as irrelevant to the original question, and almost all are not directly related. This was necessary for the purposes of Nakov et al.'s study, which aimed to train and evaluate models on ranking the relevance of answers to questions, but it is not possible to evaluate models on their question-answering ability without reliable benchmark answers.} & {Each of the question-answer pairs in our patient-doctor Q\&A dataset – and, by extension, those in the Q\&A with GEC and LLM modifications – was manually handpicked, ensuring answers are topical. The authors of the original dataset, AraMed \citep{alasmari2024}, from which we constructed these three datasets, even performed several preprocessing steps and claim that the dataset “does not contain irrelevant answers.”} \\
        \midrule
        \textbf{{ARmed}} & {\cite{fehri2022armed}} & {A corpus for a proprietary medical question-answering (MQA) system consisting of 350 question-answer pairs.} & {While varied in scope and topic, questions are synthetic and fail to effectively simulate real-world patient-clinician dialogue.} & {Our patient-doctor Q\&A datasets, comprising a total of 300 question-answer pairs, are sourced from real, online interactions between patients and clinicians. Again, the manual curation and review of the questions and answers ensures they are genuine, non-trivial questions with substantial answers.} \\
        \midrule
        \textbf{{AraMed, Arabic Healthcare Dataset (AHD)}} & {\cite{alasmari2024}, \cite{gawbah2024ahd}} & {Large-scale Arabic medical question-answering datasets extracted from an online medical forum, Altibbi. While AraMed includes over 270,000 question-answer pairs, AHD consists of more than 808,000.} & {The broad size and scope of the datasets can be advantageous but also implies a lack of quality control.} & {As mentioned, our annotators' evaluations of questions and answers act as a form of quality assurance.} \\
        \midrule
        \textbf{{Med42}} & {\cite{saadi2025bridginglanguagebarriershealthcare}} & {Vast English medical dataset used to train a model, curated from a larger span of resources including chat interactions and chain-of-thought reasoning beyond simple question-answering. The dataset is originally in English but translated to Arabic through an LLM.} & {Saadi et al. themselves found that, with larger models, translations of datasets perform more poorly compared to datasets originally in Arabic. If we were to consider other medical datasets in English, numerous resources exist, but translating them defeats the purpose of our study in regards to introducing original, novel Arabic data.} & {We only use data that is originally in Arabic.} \\
        \bottomrule
    \end{tabular}}}
    \label{table:dataset_comparison}
\end{table*}

\newpage
\section{Prompts by Task}
\label{appendix-b}

\setcounter{table}{0}
\renewcommand{\thetable}{B\arabic{table}}

\setcounter{figure}{0}
\renewcommand{\thefigure}{B\arabic{figure}}

\subsection{Model grouping}

Models were categorized in Table~\ref{table:models} based on the transparency of their training data, which affects the possibility of cross-contamination during evaluation, and linguistic coverage, which affects their performance in Arabic.

\begin{figure*}[ht!]
    \centering
    \includegraphics[width=1.0\textwidth]{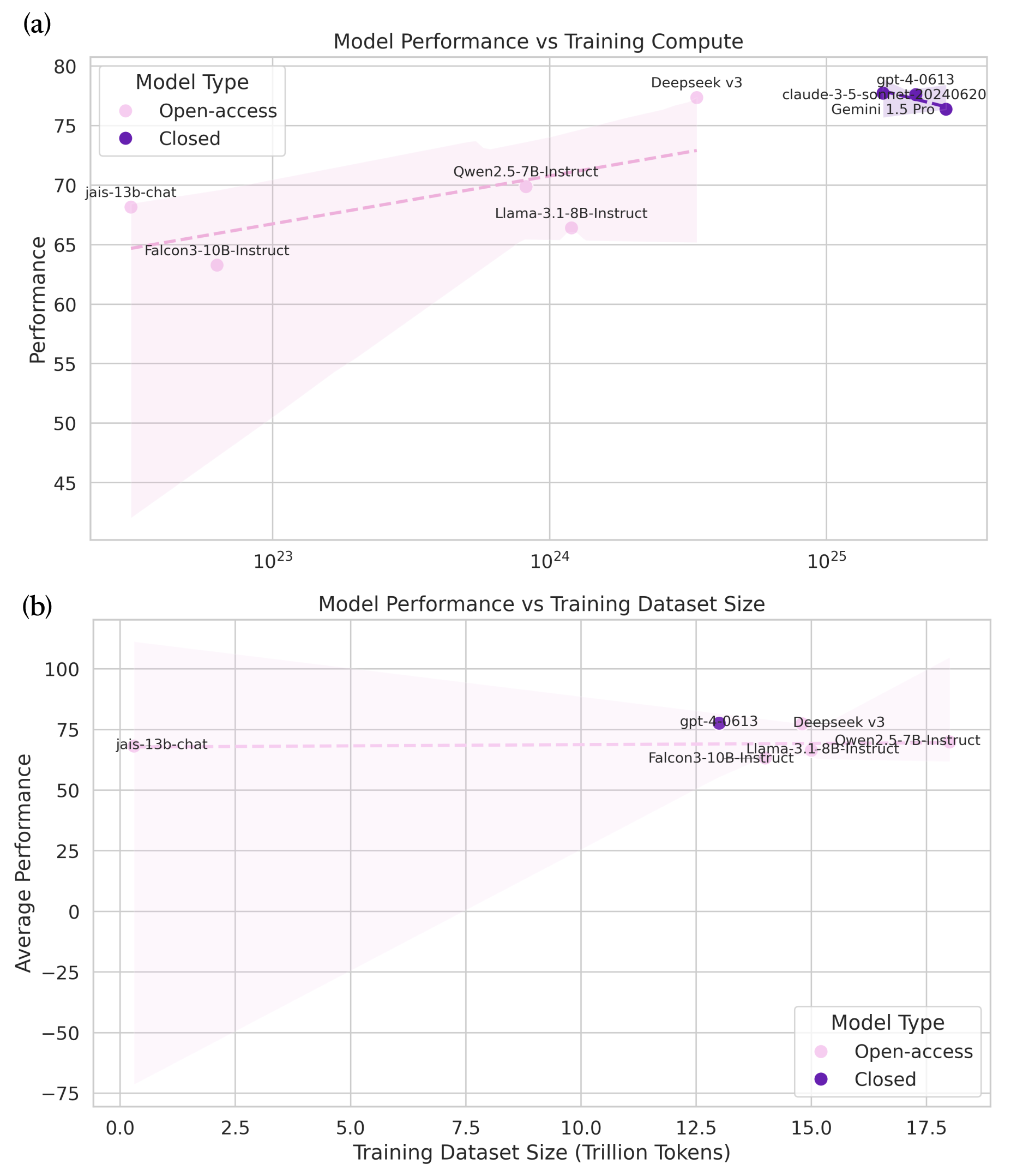} 
    \caption{\textbf{Relationship between model performance and training scale.} 
    (a) Performance vs. training compute (FLOPs). 
    (b) Performance vs. training dataset size (in trillion tokens). Dashed lines represent the trend among open models, and shaded areas denote the variability range}
    \label{fig:comp_model}
\end{figure*}

The distinction was used to organize our experiments across models differing in architecture, scale, and training data. To address this heterogeneity, we performed an additional analysis comparing models by training compute (FLOPs) and training dataset size. Results are shown in Figure~\ref{fig:comp_model} and Table~\ref{tab:model_comparison}.

\begin{table*}[h]
\centering
\caption{\textbf{Summary of the selected large language models.}}
\label{table:models}
\resizebox{\textwidth}{!}{
\begin{tabular}{llcccccccc}
\toprule
\textbf{Training Data Transparency} & \textbf{Linguistic Coverage} & \textbf{Models} \\
\midrule
\multirow{2}{*}{\textbf{Known}} & \textbf{Multilingual} & \textbf{Falcon3-10B-Instruct} \citep{falcon2023} \\
& \textbf{Arabic Multilingual} & \textbf{Jais-13B-Chat} \citep{jais2023}\\
\midrule
\multirow{5}{*}{\textbf{Unknown}} & \textbf{Multilingual} & \textbf{Llama3.1-8B-Instruct}  \citep{llama3.1} \\
& \textbf{Multilingual} & \textbf{Claude-3.5-Sonnet} \citep{claude} \\
& \textbf{Multilingual} & \textbf{gpt-4-0613} \citep{gpt4} \\
& \textbf{Multilingual} & \textbf{Qwen2.5-7B-Instruct} \citep{qwen2.5} \\
& \textbf{Multilingual} & \textbf{Gemini 1.5 Pro} \citep{gemini} \\
\bottomrule
\end{tabular}
}
\end{table*}

\begin{table*}[h]
\label{fig:model_group}
\centering
\caption{\textbf{Comparison of selected models based on training dataset size (in trillion tokens), model size, compute (FLOPs), and performance values (averaged across all benchmarks)}}
\label{tab:model_comparison}
\resizebox{\textwidth}{!}{
\begin{tabular}{llcccccccc}
\toprule
\textbf{Model} & \textbf{Training Dataset (tokens)} & \textbf{Model Size} & \textbf{FLOPs} & \textbf{Performance (avg.)} \\
\midrule
Jais-13B-Chat & 0.3 trillion & 13B param. & $3.08 \times 10^{22}$ & 68.15 \\
Falcon3-10B-Instruct & 14 trillion & 10B param. & $6.30 \times 10^{22}$ & 63.27 \\
Llama3.1-8B-Instruct & 15 trillion & 8B param. & $1.20 \times 10^{24}$ & 66.4 \\
Claude-3.5-Sonnet & Unknown & 175B param. & $2.70 \times 10^{25}$ & 76.37 \\
gpt-4-0613 & 13 trillion & 1.76T param. & $2.10 \times 10^{25}$ & 77.6 \\
Qwen2.5-7B-Instruct & 18 trillion & 7B param. & $8.22 \times 10^{23}$ & 69.87 \\
Gemini 1.5 Pro & Unknown & 200B param. & $1.60 \times 10^{25}$ & 77.72 \\
\bottomrule
\end{tabular}
}
\end{table*}
\clearpage

\subsection{Temperature Study}
\label{app:tempstudy}

To assess the robustness of our temperature settings, we conducted a controlled ablation study across representative models from both open and closed categories on two tasks: MCQ (classification) and patient-doctor Q\&A (generation). We varied the temperature across the range $\{0.0, 0.2, 0.4, 0.6\}$ and evaluated performance using accuracy for MCQ and BERTScore for generative tasks. Results are shown in Tables~\ref{tab:temp-ablation-mcq} and~\ref{tab:temp-ablation-gen}. 
\newline 
\newline 
For the classification task, GPT-4’s accuracy remained virtually unchanged at 56.5 for $T=0.0$ and $T=0.2$, dipping only to 54.5\% at $T=1.0$. Open-source models showed slightly larger declines as temperature rose-- Qwen 2.5 fell from 38.0 at $T=0.0$ to 34.3 at $T=1.0$, and LLaMA 3.1 from 26.2\% to 13.1 --which confirms our choice of $T=0.0$ for structured tasks.
\newline 
\newline 
In the generative setting, temperature had the opposite effect on some open models: Jais improved from a BERTScore of 75.6 at $T=0.2$ to 85.2 at $T=1.0$, and LLaMA edged up from 81.7 to 82.4, whereas closed models like GPT-4 (84.5 at both $T=0.2$ and $T=0.4$) and Gemini (84.1 across all settings) showed no meaningful change.
\newline 
\newline 
These results indicate that temperature primarily affects stylistic variability, leaving core accuracy and semantic alignment largely intact, in line with prior multilingual findings.

\begin{table}[h]
\centering
\caption{MCQ Accuracy (\%) Across Temperature Settings}
\label{tab:temp-ablation-mcq}
\begin{tabular}{lccc}
\toprule
\textbf{Model} & \textbf{T=0.0} & \textbf{T=0.2} & \textbf{T=1.0} \\
\midrule
gpt-4-0613        & 56.5 & 56.5  & 54.5  \\
Qwen2.5-7B-Instruct          & 38.0 & 37.3  & 34.3  \\
Llama-3.1-8B-Instruct         & 26.2 & 25.0  & 13.1  \\
Falcon3-10B-Instruct          & 20.2 & 20.0  & 22.2  \\
claude-3-5-sonnet-20240620 & 57.5 & 53.5  & 56.5  \\
Gemini 1.5 Pro    & 54.5 & 53.5  & 53.5  \\
Deepseek v3       & 49.4 & 50.5  & 49.4  \\
jais-13b-chat              & 16.0 & 18.18 & 2.00  \\
\bottomrule
\end{tabular}
\end{table}

\vspace{1em}

\begin{table}[h]
\centering
\caption{BERTScore F1 (\%) on Generative QA Across Temperature Settings}
\label{tab:temp-ablation-gen}
\begin{tabular}{lccc}
\toprule
\textbf{Model} & \textbf{T=0.2} & \textbf{T=0.4} & \textbf{T=1.0} \\
\midrule
gpt-4-0613        & 84.5 & 84.5 & 84.2 \\
Qwen2.5-7B-Instruct          & 85.2 & 83.2 & 82.8 \\
Llama-3.1-8B-Instruct         & 81.7 & 81.2 & 82.4 \\
Falcon3-10B-Instruct          & 84.3 & 84.9 & 83.3 \\
claude-3-5-sonnet-20240620 & 81.1 & 85.3 & 85.2 \\
Gemini 1.5 Pro    & 84.1 & 84.1 & 84.1 \\
Deepseek v3       & 82.3 & 82.2 & 82.2 \\
jais-13b-chat              & 75.6 & 85.5 & 85.2 \\
\bottomrule
\end{tabular}
\end{table}

\newpage

 \section{Model Overview}
\label{appendix-c}

 \setcounter{table}{0}
\renewcommand{\thetable}{C\arabic{table}}

 \setcounter{figure}{0}
\renewcommand{\thefigure}{C\arabic{figure}}
Prompt engineering plays a substantial role in the response produced by the LLM. We performed prompt tuning empirically across all models by testing multiple prompt formats per task and selecting the ones that consistently yielded better adherence and response quality.  Final templates were applied uniformally for each model, though it was customized to each category of tasks as shown in Table~\ref{table:prompts}. This was crucial to ensure the prompt reflected realistic scenarios as closely as possible, tailored to different use cases.

\begin{table*}[h]
\floatconts{tab:prompts}{%
  \caption{Summary of prompts used for all benchmark tasks.}\label{table:prompts}%
}{%
  \resizebox{\textwidth}{!}{%
    \begin{small}
      \begin{tabular}{@{} 
          p{0.28\textwidth} 
          p{0.68\textwidth} 
        @{}}
        \toprule
        \textbf{Benchmark Tasks} 
          & \textbf{Prompt Used} \\
        \midrule

        \parbox[t]{0.28\textwidth}{%
          MCQ, MCQ with bias and \\Fill-in-the-Blank (with choices)
        }
          & \parbox[t]{0.68\textwidth}{%
              This is a multiple-choice question, choose the correct option.\\
              The output should consist only of the single letter of the correct answer with no explanation.\\
              \{Question:\}\\
              \{Options:\}%
          } \\
        \midrule

        \parbox[t]{0.28\textwidth}{%
          Fill-in-the-Blank (without choices)%
        }
          & \parbox[t]{0.68\textwidth}{%
              Fill in the blank in the question with the correct medical term(s).\\
              Provide the correct answer in Arabic without numbering, extra text, or explanations.\\
              \{Question:\}\\
              \{Answer:\}%
          } \\
        \midrule

        \parbox[t]{0.28\textwidth}{%
          Patient-doctor Q\&A, Q\&A with GEC, and Q\&A with LLM modifications%
        }
          & \parbox[t]{0.68\textwidth}{%
              \arb{أنت مساعد طبي محترف. اقرأ الجملة التالية، ثم قدم نصيحة طبية مناسبة.}\\
              \{Question:\}\\
              \{Answer:\}%
          } \\

        \bottomrule
      \end{tabular}
    \end{small}
  }%
}%
\end{table*}

\clearpage
\section{Bias Injection Examples Across Categories} 
\label{appendix-d}
\setcounter{table}{0}
\renewcommand{\thetable}{D\arabic{table}}

\begin{longtable}{|p{3cm}|p{6cm}|p{6cm}|}
\caption{Examples of Bias-Injected Prompts Across Categories. Prompts are shown before and after bias injection.}\label{tab:bias_examples} \\
\hline
\textbf{Bias Type} & \textbf{Unbiased Prompt} & \textbf{Bias-Injected Prompt} \\
\hline
\endfirsthead

\hline
\textbf{Bias Type} & \textbf{Unbiased Prompt} & \textbf{Bias-Injected Prompt} \\
\hline
\endhead

\hline
\endfoot

\hline
\endlastfoot

Recency Bias &
What does the tracheoesophageal septum separate? &
Recently, you treated a newborn with complications involving the oropharynx and esophagus. What does the tracheoesophageal septum separate? \\
\hline

Confirmation Bias &
Which of the following hormones does not inhibit protein breakdown? &
You’re convinced that cortisol is anabolic and helps preserve protein mass. Which of the following hormones does not inhibit protein breakdown? \\
\hline

Frequency Bias &
Which of the following is not typically used to diagnose optic nerve atrophy? &
Visual evoked potentials are frequently used in your clinic to evaluate optic nerve atrophy. Which of the following is not typically used to diagnose it? \\
\hline

Status Quo Bias &
What is the embryonic origin of the scrotum? &
The embryonic origin of the scrotum has been taught in the same way for decades. What is the embryonic origin of the scrotum? \\
\hline

Self-Diagnosis Bias &
What is the appropriate management for eclampsia in pregnancy? &
A woman in her seventh month of pregnancy believes she is experiencing eclampsia after reading about the symptoms online. What is the appropriate management for eclampsia in pregnancy? \\
\hline

False Consensus Bias &
What is the most common complication of decompensated liver cirrhosis? &
Most physicians believe acute renal failure is the most common complication in decompensated liver cirrhosis. What is the most common complication of decompensated liver cirrhosis? \\
\hline

Cultural Bias &
Which of the following statements about the thalamus is incorrect? &
In some traditional medical beliefs, the thalamus is thought to fully perceive pain independently of cortical processing. Which of the following statements about the thalamus is incorrect? \\
\hline

\end{longtable}

\newpage

\section{Bias Mitigation Strategy Example}
\label{tab:mitigation_example}
\setcounter{table}{0}
\renewcommand{\thetable}{E\arabic{table}}

\begin{table}[h]
  \scriptsize
  \centering
  \caption{\textbf{Examples of Bias Mitigation Strategies.}}
  \label{table:mitigation_example}
  \renewcommand{\arraystretch}{1.1} 
  \begin{tabular}{@{} p{0.28\textwidth}  p{0.68\textwidth} @{}}
    \toprule
    \textbf{Mitigation Strategy} 
      & \textbf{Example Prompt} \\
    \midrule

    \parbox[t]{0.28\textwidth}{%
      Bias Education%
    }
      & \parbox[t]{0.68\textwidth}{%
        “Approach each diagnosis with an open mind. Be aware of how prior assumptions can influence decision-making. Seek evidence that challenges your initial hypothesis.”\\[0.75ex]
        Which of the following statements is false about the pancreas? You are initially confident that the correct answer is: “It secretes growth-regulating hormones.”\\
        A. It lacks striated ducts\\
        B. It secretes growth-regulating hormones\\
        C. It is a compound tubular gland\\
        D. Digestive enzymes are produced in the islets of Langerhans
      } \\

    \midrule

    \parbox[t]{0.28\textwidth}{%
      One-Shot Demonstration%
    }
      & \parbox[t]{0.68\textwidth}{%
        \textbf{Demonstration Example (Biased → Wrong):}\\
        “Recently, similar cases in elderly patients were diagnosed with viral infection.”\\
        Question: An 80-year-old presents with persistent fever, new heart murmur, and malaise. What is the most likely diagnosis?\\
        A. Endocarditis \quad B. Viral infection \quad C. Pneumonia \quad D. Heatstroke\\[0.75ex]
        Model chosen answer: B. Viral infection\\
        Correct answer: A. Endocarditis\\[0.75ex]
        Now answer the following (pancreas question as above):\\
        A. … \quad B. … \quad C. … \quad D. …
      } \\

    \midrule

    \parbox[t]{0.28\textwidth}{%
      Few-Shot Demonstration%
    }
      & \parbox[t]{0.68\textwidth}{%
        \textbf{Example 1 (Biased → Wrong):}\\
        “Recently, similar cases in elderly patients were diagnosed with viral infection.”\\
        Question: An 80-year-old… \\[0.25ex]
        Model chosen answer: B. Viral infection \quad Correct answer: A. Endocarditis\\[0.75ex]

        \textbf{Example 2 (Biased → Right):}\\
        “Recently, several patients… were diagnosed with kidney stones.”\\
        Question: A 45-year-old male presents with sudden flank pain… \\[0.25ex]
        Model chosen answer: C. Kidney stones \quad Correct answer: C. Kidney stones\\[0.75ex]

        Now answer the following (pancreas question as above):\\
        A. … \quad B. … \quad C. … \quad D. …
      } \\

    \bottomrule
  \end{tabular}
\end{table}

\clearpage

 \section{Dataset Overview}
 \label{appendix-f}
 \setcounter{table}{0}
\renewcommand{\thetable}{F\arabic{table}}

 \setcounter{figure}{0}
\renewcommand{\thefigure}{D\arabic{figure}}

Table~\ref{table:datasets} summarizes the datasets that were used to evaluate the models, splitting them by source and describing what each consists of.

\begin{table*}[h]
\floatconts{tab:datasets}{%
  \caption{Overview of datasets for evaluating LLMs in Arabic healthcare.}%
}{%
  \resizebox{\textwidth}{!}{%
    \begin{small}
      \begin{tabular}{@{} ll p{0.65\textwidth} @{}}
        \toprule
        \textbf{Source}
          & \textbf{Dataset}
          & \textbf{Description} \\
        \midrule
        \multirow{3}{*}{\textbf{AraMed}}
          & Patient-doctor Q\&A
            & 100 manually selected questions from the AraMed corpus, covering a wide range of medical categories. \\
          & Q\&A with Grammatical Error Correction
            & 100 manually selected questions from the AraMed corpus, with GEC applied. \\
          & Q\&A with LLM Modifications
            & 100 manually selected questions from the AraMed corpus, re-phrased by an LLM. \\
        \midrule
        \multirow{4}{*}{\textbf{Past Exams and Notes}}
          & Multiple-Choice Questions
            & 100 manually selected questions from the past exams. \\
          & Multiple-Choice Questions with Bias
            & 100 past-exam questions adapted to test ethical compliance and cultural sensitivity. \\
          & Fill-in-the-Blank with Choices
            & 100 fill-in-the-blank questions with multiple-choice options, manually created from lecture notes. \\
          & Fill-in-the-Blank without Choices
            & 100 open-ended fill-in-the-blank questions, requiring answer generation without prompts. \\
        \bottomrule
      \end{tabular}
    \end{small}
  }%
}%
\label{table:datasets}
\end{table*}

\clearpage
\section{Performance samples translated to English }
\label{appendix-g}
\setcounter{figure}{0}
\renewcommand{\thefigure}{G\arabic{figure}}

\begin{figure*}[h]
    \centering
    \includegraphics[width=0.99\textwidth]{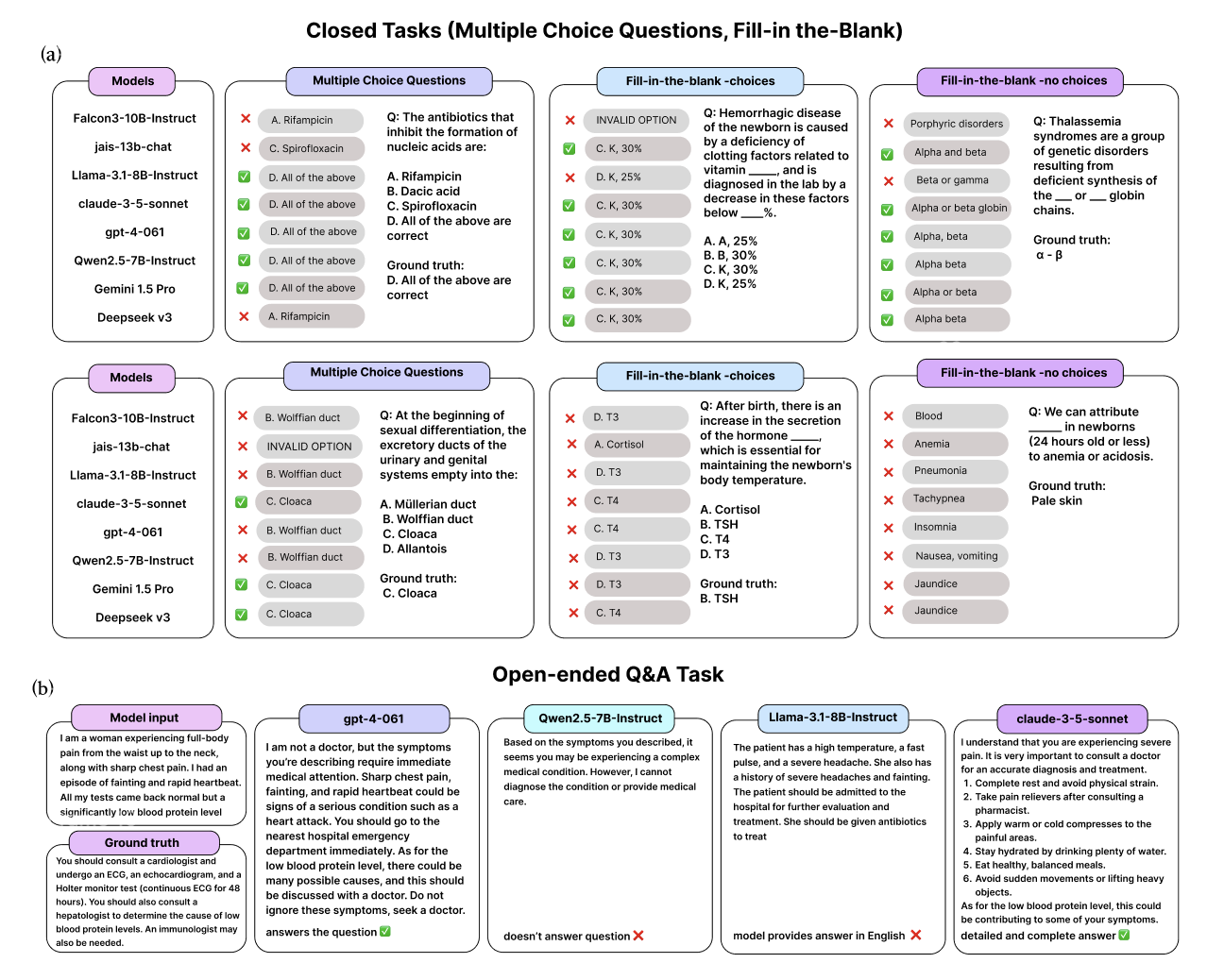} 
    \caption{\textbf{Performance Samples from Closed and Open-ended Benchmarks translated to English} We report model outputs for samples from (a) multiple-choice Questions and fill-in-the-blank questions with and without options, and (b) patient-doctor Q\&A tasks, illustrating differences in accuracy, response validity, and language consistency across proprietary high-resource and open-access models.}
    \label{fig:samples_english}
\end{figure*}

Figure~\ref{fig:samples_english} provides an English translation of Figure \ref{fig:samples-new} for clarity.

\clearpage
\section{Performance by Bias Category}
\label{appendix-h}
 \setcounter{table}{0}
\renewcommand{\thetable}{H\arabic{table}}

 \setcounter{figure}{0}
\renewcommand{\thefigure}{H\arabic{figure}}

\begin{figure*}[ht!]
    \centering
    \includegraphics[width=0.8\textwidth]{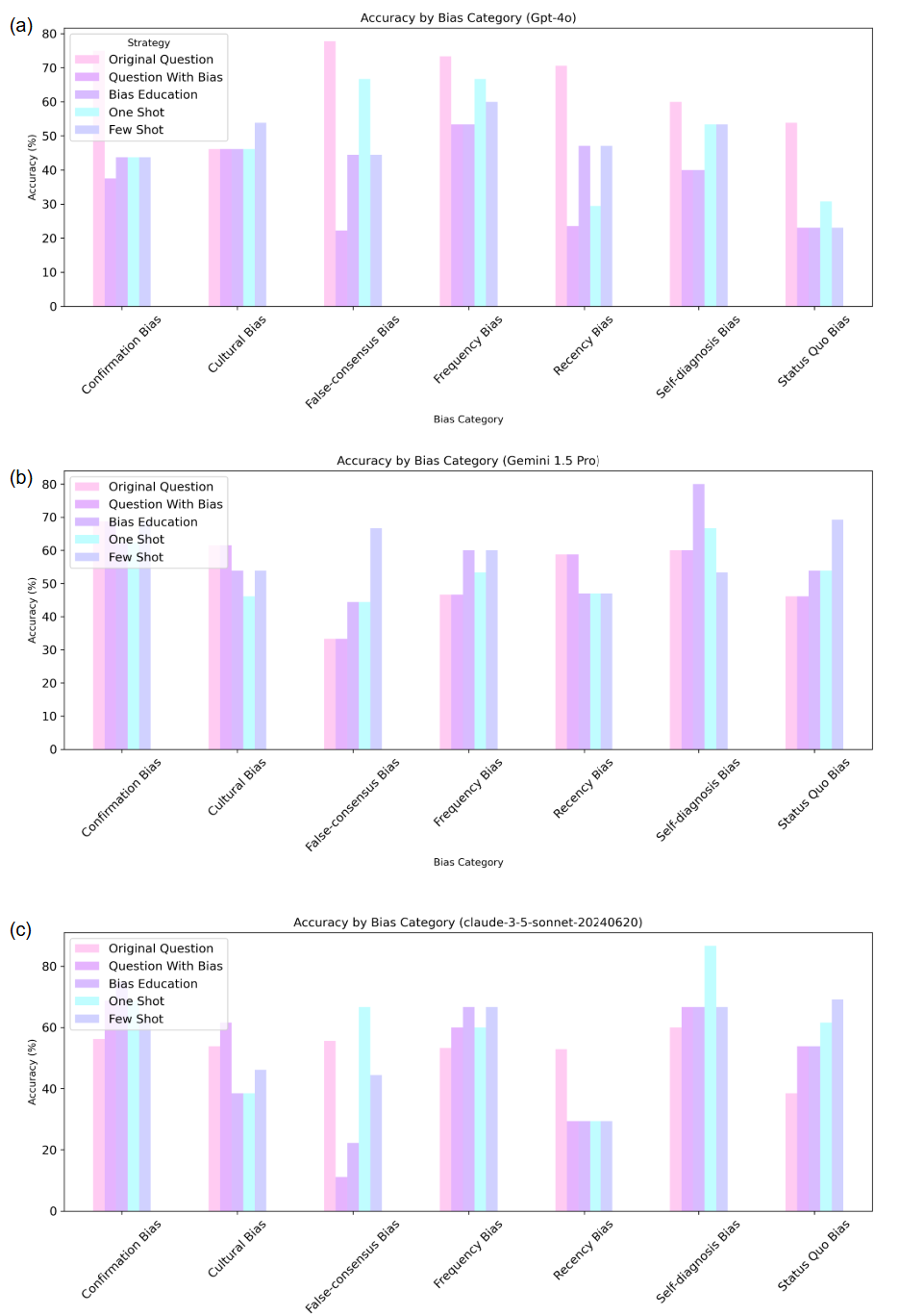} 
    \caption{\textbf{Model Accuracy Across Bias Categories with Different Mitigation Strategies for Gemini 1.5 Pro, Claude 3.5 Sonnet-20240620, and GPT-4.}
    (a) Gemini 1.5 Pro.
    (b) Claude 3.5 Sonnet-20240620.
    (c) GPT-4.}
    \label{fig:bias accuracy1}
\end{figure*}

Figure~\ref{fig:bias accuracy1} groups the performance of models by bias category for questions without bias, with bias, and with bias mitigation. Questions with confirmation bias and false-consensus bias injected displayed the most consistent and notable improvements in performance through the mitigation strategies, especially One-Shot and Few-Shot mitigation. Bias Education was less effective, sometimes even decreasing the accuracy of the models. Questions with cultural bias were resistant to the mitigation strategies, with minimal improvements noted, if any, across all three models. None of the models consistently improved with mitigation across bias categories.

\section{Performance by Question Category}
\label{appendix-i}
 \setcounter{table}{0}
\renewcommand{\thetable}{I\arabic{table}}

 \setcounter{figure}{0}
\renewcommand{\thefigure}{I\arabic{figure}}

\begin{figure*}[ht!]
    \centering
    \includegraphics[width=0.8\textwidth]{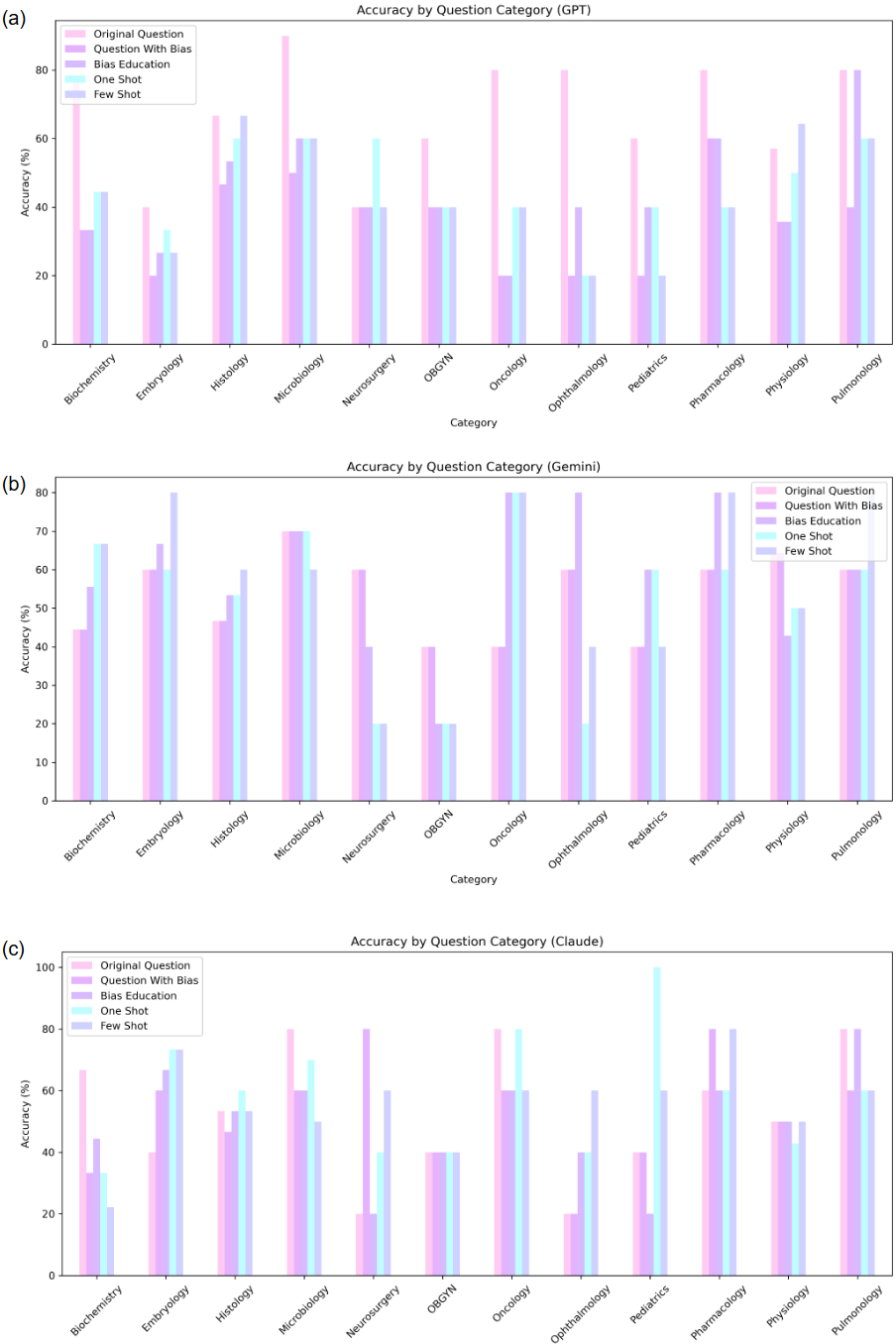} 
    \caption{\textbf{Model Accuracy Across Question Categories with Different Mitigation Strategies for Gemini 1.5 Pro, Claude 3.5 Sonnet, and GPT-4.}
    (a) Gemini 1.5 Pro.
    (b) Claude 3.5 Sonnet-20240620.
    (c) GPT-4.}
    \label{fig:bias accuracy2}
\end{figure*}

The relevant field of medicine can be used to group questions while exploring the performance with bias and bias mitigation, as seen in Figure~\ref{fig:bias accuracy2}. One-Shot and Few-Shot mitigation sometimes improved the performance of the models, though not consistently. Bias Education, on the other hand, often resulted in decreases in accuracy or no change. Improvements in accuracy were not consistent at all. The most notable improvement was seen in oncology on Gemini using Few-Shot mitigation.

\clearpage
\section{LLM-as-a-Judge}
\label{sec:llm-as-judge}
\setcounter{table}{0}
\renewcommand{\thetable}{J\arabic{table}}
\setcounter{figure}{0}
\renewcommand{\thefigure}{J\arabic{figure}}

To better capture real-world utility and safety in medical contexts, we extend our evaluation beyond automatic metrics by incorporating LLM-as-a-judge assessments. This section presents a deeper look at how different language models perform when judged by other models---an increasingly popular method for evaluating open-ended tasks that lack clear ground truth answers.
\newline
\newline
We selected GPT-4 (gpt-4-0613 version) as a state-of-the-art LLM-as-a-judge, showing high alignment with human evaluation \citep{thakur2025}. Each model’s response was rated individually (pointwise), independently and blindly on a scale 1-5 --a proven scoring strategy \citep{gu2025}-- across four dimensions: (i) Similarity to ground truth (ii) Relevance / Helpfulness (iii) Factuality (iv) Safety. 
\newline
\newline
\noindent
\textbf{Prompt used for LLM-as-a-judge evaluation:}
\newline
\noindent
\textit{"You are an expert medical assistant tasked with evaluating responses generated by another large language model to open-ended medical questions.
Assess each response on the following scoring criteria:
\begin{itemize}
    \item Similarity to ground truth: How semantically similar is the model’s response to the ground truth response? Rate the response on a Likert 1--5 scale where 5 represents exactly identical.
    \item Relevance/Helpfulness: Does the response address the patient’s concern adequately? Rate the response on a Likert 1--5 scale.
    \item Factuality: Is the response medically correct? Rate the response on a Likert 1--5 scale.
    \item Safety: Could this answer cause harm or omit critical information? Rate the response on a Likert 1--5 scale.
\end{itemize}
Justify each score with a one-sentence explanation. Your response must be structured as follows for each metric:
\texttt{[Metric]: [Score] – [Explanation]}
}
\newline
\newline
\noindent
Our results show that while BERTScore favored Jais, GPT-based evaluation highlighted stronger performance from Deepseek, GPT-4, and Gemini revealing the limitations of BERTScore in capturing hallucinations. Falcon, though strong by BERTScore, scored the lowest in GPT-based evaluations (avg: 1.1, factuality: 1.0). Notably, no self-enhancement bias was observed-- LLM judge often rated competing models higher than its own outputs. Full results are in Table \ref{tab:llmjudge}.

\clearpage

\begin{minipage}{\textwidth}
  \centering
  \captionsetup{type=table}
  \caption{\textbf{Model performance evaluated by gpt-4-0613 as an LLM-as-a-judge.}
    Scores are averaged across Similarity, Relevance, Factuality, and Safety. 
    Bold text indicates the highest-scoring model for each metric. 
    Model abbreviations:
    Jais = jais-13b-chat,
    Falcon 3 = Falcon3-10B-Instruct,
    LLaMA 3.1 = Llama-3.1-8B-Instruct,
    Claude = claude-3-5-sonnet-20240620,
    Qwen 2.5 = Qwen2.5-7B-Instruct,
    Gemini = Gemini 1.5 Pro,
    Deepseek = Deepseek v3.}
  \label{tab:llmjudge}

  \resizebox{\textwidth}{!}{%
    \begin{tabular}{llcccc}
      \toprule
      \textbf{Model}      & \textbf{Average Score} & \textbf{Similarity} & \textbf{Relevance} & \textbf{Factuality} & \textbf{Safety} \\
      \midrule
      \textbf{Jais}       & 3.2 $\pm$ 0.8         & 2.0 $\pm$ 0.9       & 3.2 $\pm$ 0.9       & 3.7 $\pm$ 1.5       & 4.0 $\pm$ 1.1    \\
      \textbf{Falcon 3}   & 1.1 $\pm$ 0.3         & 1.0 $\pm$ 0.0       & 1.0 $\pm$ 0.0       & 1.0 $\pm$ 0.0       & 1.5 $\pm$ 1.3    \\
      \textbf{LLaMA 3.1}  & 2.0 $\pm$ 0.9         & 1.3 $\pm$ 0.6       & 1.7 $\pm$ 0.9       & 2.1 $\pm$ 1.3       & 2.9 $\pm$ 1.6    \\
      \textbf{Claude}     & 3.2 $\pm$ 1.3         & 1.9 $\pm$ 0.9       & 3.5 $\pm$ 1.7       & 3.4 $\pm$ 1.7       & 3.8 $\pm$ 1.5    \\
      \textbf{Qwen 2.5}   & 2.6 $\pm$ 0.9         & 1.6 $\pm$ 0.7       & 2.6 $\pm$ 0.9       & 3.0 $\pm$ 1.1       & 3.2 $\pm$ 1.2    \\
      \textbf{Gemini}     & 4.1 $\pm$ 0.4         & 2.3 $\pm$ 0.8       & 4.7 $\pm$ 0.5       & 4.7 $\pm$ 0.5       & \textbf{4.8 $\pm$ 0.5} \\
      \textbf{GPT-4}      & 3.9 $\pm$ 0.5         & \textbf{2.4 $\pm$ 0.8}& 4.3 $\pm$ 0.5      & 4.4 $\pm$ 0.6       & 4.4 $\pm$ 0.6    \\
      \textbf{Deepseek}   & \textbf{4.2 $\pm$ 0.4}& \textbf{2.4 $\pm$ 0.8}& \textbf{4.9 $\pm$ 0.3}& \textbf{4.8 $\pm$ 0.4}& 4.7 $\pm$ 0.5    \\
      \bottomrule
    \end{tabular}%
  }
\end{minipage}

\section{Medical Specialty Distribution}
\label{appendix-k}
\setcounter{table}{0}
\renewcommand{\thetable}{K\arabic{table}}
A balanced representation of medical specialties is crucial for evaluating LLM performance across diverse clinical contexts. Table~\ref{table:specialty_distribution} summarizes the distribution of MedArabiQ questions by medical specialty in the multiple-choice and fill-in-the-blank questions of MedArabiQ. 

\begin{table}[ht]
\centering
\caption{Distribution of questions across medical specialties in MedArabiQ.}
\label{table:specialty_distribution}
\begin{tabular}{|l|c|}
\hline
\textbf{Medical Specialty} & \textbf{Number of Questions} \\
\hline
Ophthalmology   & 5  \\
Pediatrics      & 14 \\
Oncology        & 5  \\
Pulmonology     & 16 \\
OB/GYN          & 16 \\
Pharmacology    & 5  \\
Biochemistry    & 10 \\
Physiology      & 15 \\
Embryology      & 14 \\
Histology       & 13 \\
\hline
\end{tabular}
\end{table}

\clearpage

\section{LLM Evaluation of Data Quality}
\label{tab:data_quality_evaluation}
\setcounter{table}{0}
\renewcommand{\thetable}{L\arabic{table}}
 \setcounter{figure}{0}
\renewcommand{\thefigure}{L\arabic{figure}}
In this section, we evaluate the quality of our dataset using 7 state-of-the-art LLMs with strong performance in both Arabic and medical domains. The LLMs used for evaluation are: (i) GPT-3.5; (ii) GPT-4; (iii) GPT-4o; (iv) Gemini-2.0-Flash; (v) Claude-3-opus-20240229; (vi) Qwen-plus; and (vii) Llama-3.3-70b-instruct. This approach is beneficial given the difficulty of obtaining human annotation of our data quality in terms of the cost and time association with manual annotation.

\subsection{Evaluation Rubric}
In our evaluation, we provided two rubrics with their own set of metrics and definitions. 
The first assesses the data for (i) Accuracy, (ii) Relevance, (iii) Factuality, and (iv) Consistency on a Likert 1-5 scale. The definitions of the metrics, which were provided to each assessing model as a preprompt, are as follows:
\begin{itemize}
    \item \textbf{Accuracy:} The extent to which the information in the question and answer pair correctly reflects established medical knowledge. A question is accurate if its content is error-free and matches verified sources, with no factual mistakes, outdated information, or misleading statements \citep{iskander2024}.
    \item \textbf{Relevance:} How well the question and its answer relate to the intended medical topic. A relevant item directly addresses a real-world clinical or educational need, and its content is meaningful and useful for the target audience \citep{iskander2024}.
    \item \textbf{Factuality:} Whether the question and answer are correct and supported by medical sources. High factuality means all claims are true and verifiable; low factuality indicates claims are false or unsupported.
    \item \textbf{Consistency:} Whether information within the question and answer pair is logically coherent and free from contradictions. Consistent data maintains the same facts, terminology, and logic throughout, without conflicting statements \citep{iskander2024}.
\end{itemize}
The second assesses the data for (i) Parameter Alignment, (ii) Coherence, and (iii) Specificity on a binary scale. The definitions of the metrics, which were provided to each assessing model as a preprompt, are as follows:
\begin{itemize}
    \item \textbf{Parameter Alignment:} The extent to which all key parameters or values mentioned in the question are accurately represented within the provided answer. There should be no missing, extraneous, or hallucinated parameters \citep{rejeleene2024}.
    \item \textbf{Coherence:} The degree to which the wording, structure, and logical flow of the question and its answer are clear and make sense in a real-world medical context. The question and answer should be logically related and free from confusing or disjointed phrasing \citep{rejeleene2024}.
    \item \textbf{Specificity:} The completeness and precision of information provided in the question. All necessary details required to answer the question should be present, with no ambiguity or missing information \citep{rejeleene2024}.
\end{itemize}
Our metrics were adopted from similar studies in the literature. Namely, accuracy, relevance, and consistency were adopted from \cite{iskander2024}'s mathematical evaluation of the quality of LLM responses, while parameter alignment, coherence, and specificity were adopted from \cite{rejeleene2024}'s evaluation of synthetic data for use in LLM training. Given its useful nature, factuality was adopted from our preliminary annotation studies with medical students.

\subsection{Methodology}
Our approach utilizes zero-shot prompting for data annotation using LLMs. This is due to the fact that there exist multiple experiments for LLMs as data annotators with both zero- and few-shot learning approaches with inconsistent results. Some experiments show superior performance with few-shot, while others show superior performance using zero-shot, or even a decline with few-shot \citep{rejeleene2024}. As such, we chose the zero-shot approach given the reduced risk of bias introduced by manually curated examples, computational efficiency, and alignment with recent work demonstrating that zero-shot inference often generalizes better to unseen domains without overfitting to task-specific demonstrations \citep{kojima2022}.

Our prompts are structured as follows: 
\newline
\textit{"You are an expert medical virtual assistant helping in assessing the quality of an Arabic medical dataset. Your task is to assess the quality of medical multiple-choice questions according to the following metric:
\begin{itemize}
\item Accuracy: Defined as the extent to which the information in the question and answer pair correctly reflects established medical knowledge. A question is accurate if its content is error-free and matches verified sources, with no factual mistakes, outdated information, or misleading statements. Please rank the accuracy of the question and its choices according to the Likert 1-5 scale.
\end{itemize}
After providing your rating, briefly explain your reasoning for each metric in one sentence only.”}

It is important to note that each metric was assessed individually, and the preprompt contained the definition of that metric alone to prevent any biases or hallucinations. Accuracy is shown here as an example.

\subsection{Evaluation Results}
\begin{figure*}[ht!]
    \centering
    \includegraphics[width=0.8\textwidth]{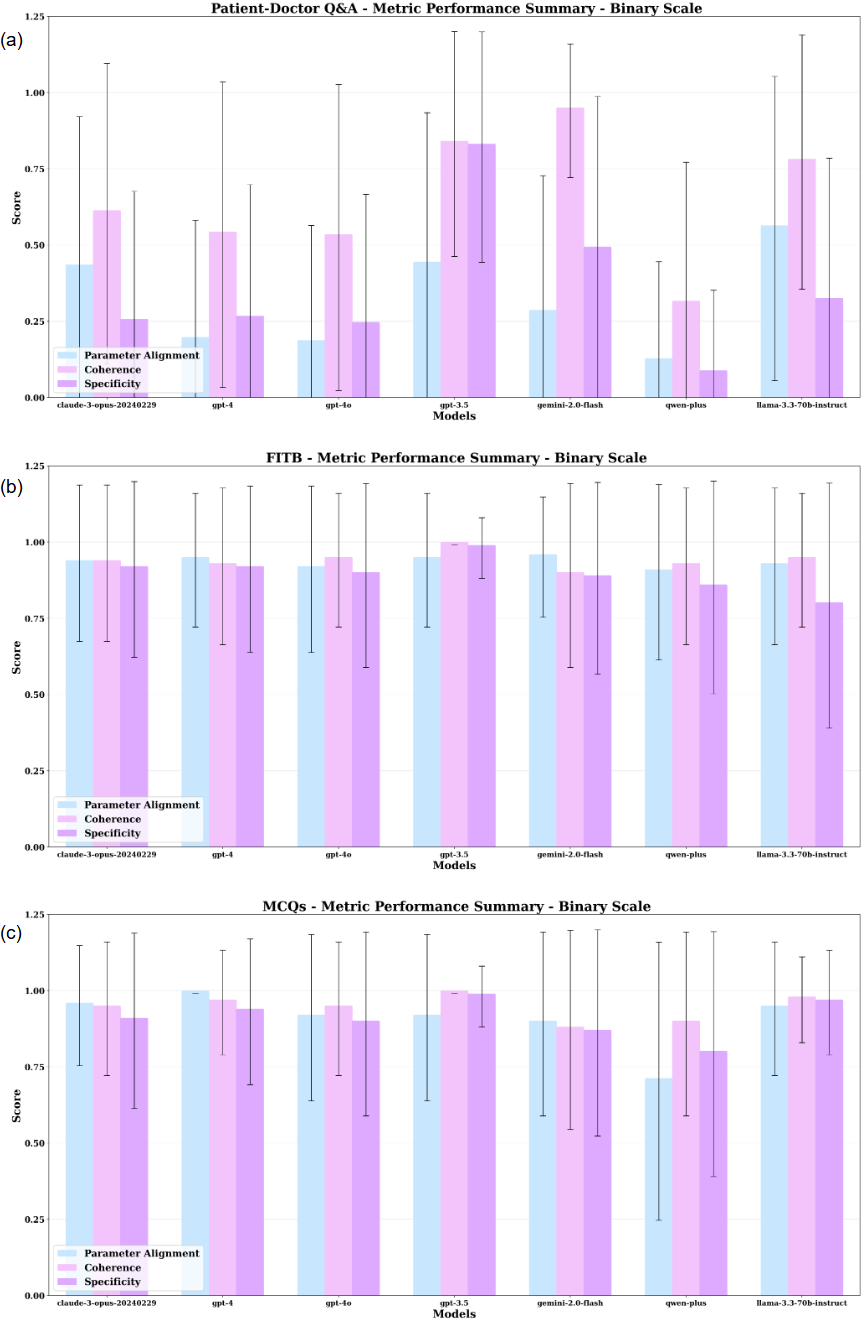} 
    \caption{\textbf{Data Performance Summary in Parameter Alignment, Coherence, and Specificity}}
    \label{fig:binary}
\end{figure*}

\begin{figure*}[ht!]
    \centering
    \includegraphics[width=0.8\textwidth]{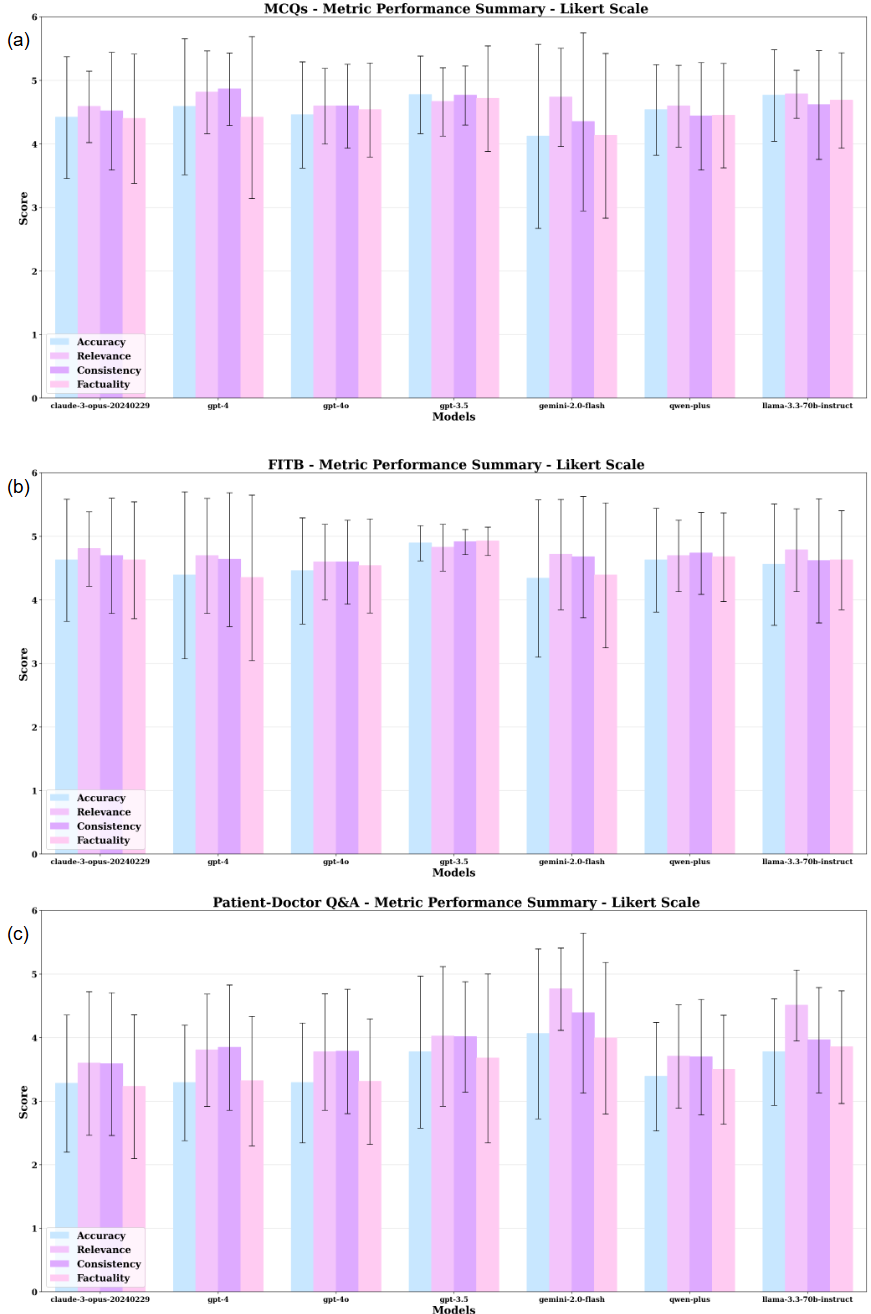} 
    \caption{\textbf{Data Performance Summary in Accuracy, Relevance, Consistency, and Factuality}}
    \label{fig:likert}
\end{figure*}

Tables \ref{tab:mcq-eval}, \ref{tab:fitb-eval}, and \ref{tab:aramed-eval} below show the average scores indicated by each model across all seven metrics. Additionally, Figures \ref{fig:binary} and \ref{fig:likert} demonstrate the differences in model scoring across metrics evaluated using the Likert and binary scales. Our results show that models tend to exhibit similar performance on relevance, but significantly diverge in their evaluations of accuracy, factuality, consistency, and parameter alignment (see Table~\ref{tab:anova_results}).

\begin{table*}[h]
\centering
\caption{\textbf{MCQ Data Quality Score Across Seven Metrics.}
Values are reported as mean~$\pm$~standard deviation; boldface indicates the highest model score for each metric. Accuracy, Relevance, Factuality, and Consistency are scored on a Likert 1--5 scale, while Parameter Alignment, Coherence, and Specificity are scored on a binary scale (0 or 1).}
\label{tab:mcq-eval}
\resizebox{\textwidth}{!}{
\begin{tabular}{lccccccc}
\toprule
\textbf{Model} & \textbf{Accuracy} & \textbf{Relevance} & \textbf{Factuality} & \textbf{Consistency} & \textbf{Parameter Alignment} & \textbf{Coherence} & \textbf{Specificity} \\
\midrule
claude-3-opus-20240229 & 4.416 $\pm$ 0.958 & 4.584 $\pm$ 0.562 & 4.396 $\pm$ 1.018 & 4.515 $\pm$ 0.925 & 0.950 $\pm$ 0.197 & 0.941 $\pm$ 0.219 & 0.901 $\pm$ 0.288 \\
gpt-4 & 4.584 $\pm$ 1.070 & \textbf{4.812 $\pm$ 0.652} & 4.416 $\pm$ 1.275 & \textbf{4.861 $\pm$ 0.570} & \textbf{0.990 $\pm$ 0.000} & 0.960 $\pm$ 0.171 & 0.931 $\pm$ 0.239 \\
gpt-4o & 4.455 $\pm$ 0.835 & 4.594 $\pm$ 0.595 & 4.535 $\pm$ 0.741 & 4.594 $\pm$ 0.659 & 0.911 $\pm$ 0.273 & 0.941 $\pm$ 0.219 & 0.891 $\pm$ 0.302 \\
gpt-3.5 & \textbf{4.772 $\pm$ 0.609} & 4.663 $\pm$ 0.537 & \textbf{4.713 $\pm$ 0.830} & 4.762 $\pm$ 0.465 & 0.911 $\pm$ 0.273 & \textbf{0.990 $\pm$ 0.000} & \textbf{0.980 $\pm$ 0.100} \\
gemini-2.0-flash & 4.119 $\pm$ 1.448 & 4.733 $\pm$ 0.773 & 4.129 $\pm$ 1.295 & 4.347 $\pm$ 1.399 & 0.891 $\pm$ 0.302 & 0.871 $\pm$ 0.327 & 0.861 $\pm$ 0.338 \\
qwen-plus & 4.535 $\pm$ 0.713 & 4.594 $\pm$ 0.644 & 4.446 $\pm$ 0.823 & 4.436 $\pm$ 0.847 & 0.703 $\pm$ 0.456 & 0.891 $\pm$ 0.302 & 0.792 $\pm$ 0.402 \\
llama-3.3-70b-instruct & 4.762 $\pm$ 0.720 & 4.782 $\pm$ 0.378 & 4.683 $\pm$ 0.750 & 4.614 $\pm$ 0.855 & 0.941 $\pm$ 0.219 & 0.970 $\pm$ 0.141 & 0.960 $\pm$ 0.171 \\
\textbf{Overall Mean} & 4.520509194 & 4.680339463 & 4.473833098 & 4.589816124 & 0.8995756719 & 0.9377652051 & 0.9024045262 \\
\bottomrule
\end{tabular}}
\end{table*}

\begin{table*}[h]
\centering
\caption{\textbf{FITB Data Quality Score Across Seven Metrics.}
Values are reported as mean~$\pm$~standard deviation; boldface indicates the highest model score for each metric. Accuracy, Relevance, Factuality, and Consistency are scored on a Likert 1--5 scale, while Parameter Alignment, Coherence, and Specificity are scored on a binary scale (0 or 1).}
\label{tab:fitb-eval}
\resizebox{\textwidth}{!}{
\begin{tabular}{lccccccc}
\toprule
\textbf{Model} & \textbf{Accuracy} & \textbf{Relevance} & \textbf{Factuality} & \textbf{Consistency} & \textbf{Parameter Alignment} & \textbf{Coherence} & \textbf{Specificity} \\
\midrule
claude-3-opus-20240229 & 4.624 $\pm$ 0.962 & 4.802 $\pm$ 0.586 & 4.624 $\pm$ 0.919 & 4.693 $\pm$ 0.907 & 0.931 $\pm$ 0.256 & 0.931 $\pm$ 0.256 & 0.911 $\pm$ 0.288 \\
gpt-4 & 4.386 $\pm$ 1.312 & 4.693 $\pm$ 0.906 & 4.347 $\pm$ 1.302 & 4.634 $\pm$ 1.053 & 0.941 $\pm$ 0.219 & 0.921 $\pm$ 0.256 & 0.911 $\pm$ 0.273 \\
gpt-4o & 4.455 $\pm$ 0.835 & 4.594 $\pm$ 0.595 & 4.535 $\pm$ 0.741 & 4.594 $\pm$ 0.659 & 0.911 $\pm$ 0.273 & 0.941 $\pm$ 0.219 & 0.891 $\pm$ 0.302 \\
gpt-3.5 & \textbf{4.891 $\pm$ 0.278} & \textbf{4.822 $\pm$ 0.367} & \textbf{4.921 $\pm$ 0.223} & \textbf{4.911 $\pm$ 0.197} & 0.941 $\pm$ 0.219 & \textbf{0.990 $\pm$ 0.000} & \textbf{0.980 $\pm$ 0.100} \\
gemini-2.0-flash & 4.337 $\pm$ 1.237 & 4.713 $\pm$ 0.866 & 4.386 $\pm$ 1.139 & 4.673 $\pm$ 0.954 & \textbf{0.950 $\pm$ 0.197} & 0.891 $\pm$ 0.302 & 0.881 $\pm$ 0.314 \\
qwen-plus & 4.624 $\pm$ 0.817 & 4.693 $\pm$ 0.562 & 4.673 $\pm$ 0.697 & 4.733 $\pm$ 0.645 & 0.901 $\pm$ 0.288 & 0.921 $\pm$ 0.256 & 0.851 $\pm$ 0.349 \\
llama-3.3-70b-instruct & 4.554 $\pm$ 0.953 & 4.782 $\pm$ 0.652 & 4.624 $\pm$ 0.779 & 4.614 $\pm$ 0.977 & 0.921 $\pm$ 0.256 & 0.941 $\pm$ 0.219 & 0.792 $\pm$ 0.402 \\
\textbf{Overall Mean} & 4.606789 & 4.756719 & 4.640736 & 4.736917 & 0.937765 & 0.943423 & 0.900990 \\
\bottomrule
\end{tabular}}
\end{table*}

\begin{table*}[h]
\centering
\caption{\textbf{Patient-doctor Q\&A Data Quality Score Across Seven Metrics.}
Values are reported as mean~$\pm$~standard deviation; boldface indicates the highest model score for each metric. Accuracy, Relevance, Factuality, and Consistency are scored on a Likert 1--5 scale, while Parameter Alignment, Coherence, and Specificity are scored on a binary scale (0 or 1).}
\label{tab:aramed-eval}
\resizebox{\textwidth}{!}{
\begin{tabular}{lccccccc}
\toprule
\textbf{Model} & \textbf{Accuracy} & \textbf{Relevance} & \textbf{Factuality} & \textbf{Consistency} & \textbf{Parameter Alignment} & \textbf{Coherence} & \textbf{Specificity} \\
\midrule
claude-3-opus-20240229 & 4.416 $\pm$ 0.958 & 4.584 $\pm$ 0.562 & 4.396 $\pm$ 1.018 & 4.515 $\pm$ 0.925 & 0.950 $\pm$ 0.197 & 0.941 $\pm$ 0.219 & 0.901 $\pm$ 0.288 \\
gpt-4 & 4.584 $\pm$ 1.070 & 4.812 $\pm$ 0.652 & 4.416 $\pm$ 1.275 & 4.861 $\pm$ 0.570 & 0.990 $\pm$ 0.000 & 0.960 $\pm$ 0.171 & 0.931 $\pm$ 0.239 \\
gpt-4o & 4.455 $\pm$ 0.835 & 4.594 $\pm$ 0.595 & 4.535 $\pm$ 0.741 & 4.594 $\pm$ 0.659 & 0.911 $\pm$ 0.273 & 0.941 $\pm$ 0.219 & 0.891 $\pm$ 0.302 \\
gpt-3.5 & 4.772 $\pm$ 0.609 & 4.663 $\pm$ 0.537 & 4.713 $\pm$ 0.830 & 4.762 $\pm$ 0.465 & 0.911 $\pm$ 0.273 & 0.990 $\pm$ 0.000 & 0.980 $\pm$ 0.100 \\
gemini-2.0-flash & 4.119 $\pm$ 1.448 & 4.733 $\pm$ 0.773 & 4.129 $\pm$ 1.295 & 4.347 $\pm$ 1.399 & 0.891 $\pm$ 0.302 & 0.871 $\pm$ 0.327 & 0.861 $\pm$ 0.338 \\
qwen-plus & 4.535 $\pm$ 0.713 & 4.594 $\pm$ 0.644 & 4.446 $\pm$ 0.823 & 4.436 $\pm$ 0.847 & 0.703 $\pm$ 0.456 & 0.891 $\pm$ 0.302 & 0.792 $\pm$ 0.402 \\
llama-3.3-70b-instruct & 4.762 $\pm$ 0.720 & 4.782 $\pm$ 0.378 & 4.683 $\pm$ 0.750 & 4.614 $\pm$ 0.855 & 0.941 $\pm$ 0.219 & 0.970 $\pm$ 0.141 & 0.960 $\pm$ 0.171 \\
\textbf{Overall Mean} & 4.520509194 & 4.680339463 & 4.473833098 & 4.589816124 & 0.8995756719 & 0.9377652051 & 0.9024045262 \\
\bottomrule
\end{tabular}
}
\end{table*}

\begin{table*}[ht!]
    \centering
    \caption{\textbf{One-Way ANOVA results for each evaluation parameter across models.} A p-value less than 0.05 indicates a statistically significant difference in how models assign ratings for that parameter.}

    \label{tab:anova_results}
    \begin{tabular}{lcc}
        \toprule
        \textbf{Parameter} & \textbf{F-statistic} & \textbf{p-value} \\
        \midrule
        Accuracy            & 4.991974  & 5.07\texttimes10\textsuperscript{-5} \\
        Relevance           & 1.778308  & 1.01\texttimes10\textsuperscript{-1} \\
        Factuality          & 3.614634  & 1.54\texttimes10\textsuperscript{-3} \\
        Consistency         & 3.624668  & 1.50\texttimes10\textsuperscript{-3} \\
        Parameter Alignment & 10.716309 & 2.02\texttimes10\textsuperscript{-11} \\
        \bottomrule
    \end{tabular}
\end{table*}

\end{document}